\documentclass[10pt,twocolumn,letterpaper]{article}

\usepackage{cvpr}              

\usepackage{algorithm}
\usepackage{algorithmic}
\usepackage{multirow}
\usepackage[dvipsnames]{xcolor}

\definecolor{cvprblue}{rgb}{0.21,0.49,0.74}
\usepackage[pagebackref,breaklinks,colorlinks,citecolor=cvprblue]{hyperref}

\def\paperID{5104} 
\def\confName{CVPR}
\def\confYear{2024}

\title{Exploiting Inter-sample and Inter-feature Relations in Dataset Distillation}

\author{Wenxiao Deng\textsuperscript{$1$}, Wenbin Li\textsuperscript{$1$}\thanks{Corresponding author}, Tianyu Ding\textsuperscript{$2$}, Lei Wang\textsuperscript{$3$},\\
Hongguang Zhang\textsuperscript{$4$}, Kuihua Huang\textsuperscript{$5$}, Jing Huo\textsuperscript{$1$}, Yang Gao\textsuperscript{$1$} \\ 
}

\date{\textsuperscript{$1$}State Key Laboratory for Novel Software Technology, Nanjing University, China \\
\textsuperscript{$2$}Microsoft Corporation, USA \quad
\textsuperscript{$3$}University of Wollongong, Australia\quad \\
\textsuperscript{$4$}Systems Engineering Institute, AMS, China\quad 
\textsuperscript{$5$}School of Systems Engineering, University of Defense Technology, China\quad}

\begin{document}
\maketitle
\begin{abstract}

Dataset distillation has emerged as a promising approach in deep learning, enabling efficient training with small synthetic datasets derived from larger real ones. Particularly, distribution matching-based distillation methods attract attention thanks to its effectiveness and low computational cost. However, these methods face two primary limitations: the dispersed feature distribution within the same class in synthetic datasets, reducing class discrimination, and an exclusive focus on mean feature consistency, lacking precision and comprehensiveness. To address these challenges, we introduce two novel constraints: a class centralization constraint and a covariance matching constraint. The class centralization constraint aims to enhance class discrimination by more closely clustering samples within classes. The covariance matching constraint seeks to achieve more accurate feature distribution matching between real and synthetic datasets through local feature covariance matrices, particularly beneficial when sample sizes are much smaller than the number of features. Experiments demonstrate notable improvements with these constraints, yielding performance boosts of up to 6.6\% on CIFAR10, 2.9\% on SVHN, 2.5\% on CIFAR100, and 2.5\% on TinyImageNet, compared to the state-of-the-art relevant methods. In addition, our method maintains robust performance in cross-architecture settings, with a maximum performance drop of 1.7\% on four architectures. Code is available at \url{https://github.com/VincenDen/IID}.

\end{abstract}

\section{Introduction}
\label{sec:intro}

Deep learning has evolved rapidly in recent years, achieving remarkable results across various fields~\cite{dl_dosovitskiy2020image,dl_huang2017densely,dl_liu2021swin,dl_redmon2016you,dl_tsai2018learning,YangZZX019,YangSZFZXY23}. Dataset distillation~\cite{wang2018dataset}, a process of distilling knowledge from a large real dataset to a much smaller synthetic dataset, has emerged as a key technique for efficient deep learning training. It has also been widely adopted in areas like neural architecture search~\cite{du2023minimizing,cui2022dc,zhao2020dataset,zhao2021dataset,zhao2023dataset}, continual learning~\cite{rosasco2021distilled,du2023minimizing,sangermano2022sample}, privacy protection~\cite{dong2022privacy,zheng2022differentially,carlini2022no}.
    \begin{figure}[t]
      \centering
      \begin{subfigure}{0.45\linewidth}
        \includegraphics[width=0.8\linewidth]{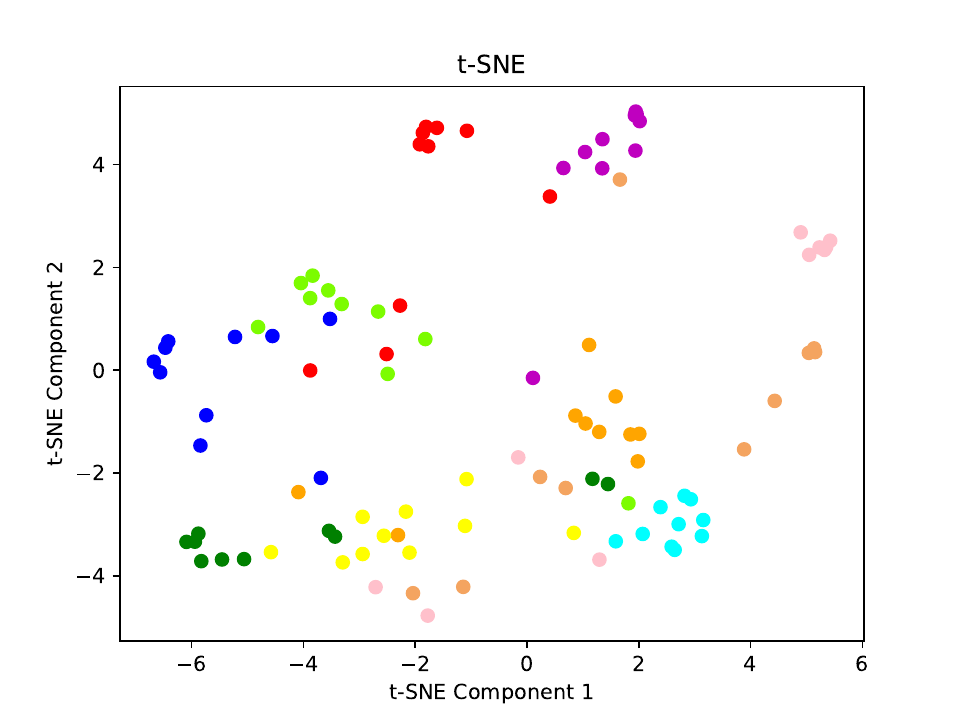}
        \caption{DM IPC=10}
        \label{fig:DM_10}
      \end{subfigure}
      \begin{subfigure}{0.45\linewidth}
        \includegraphics[width=0.8\linewidth]{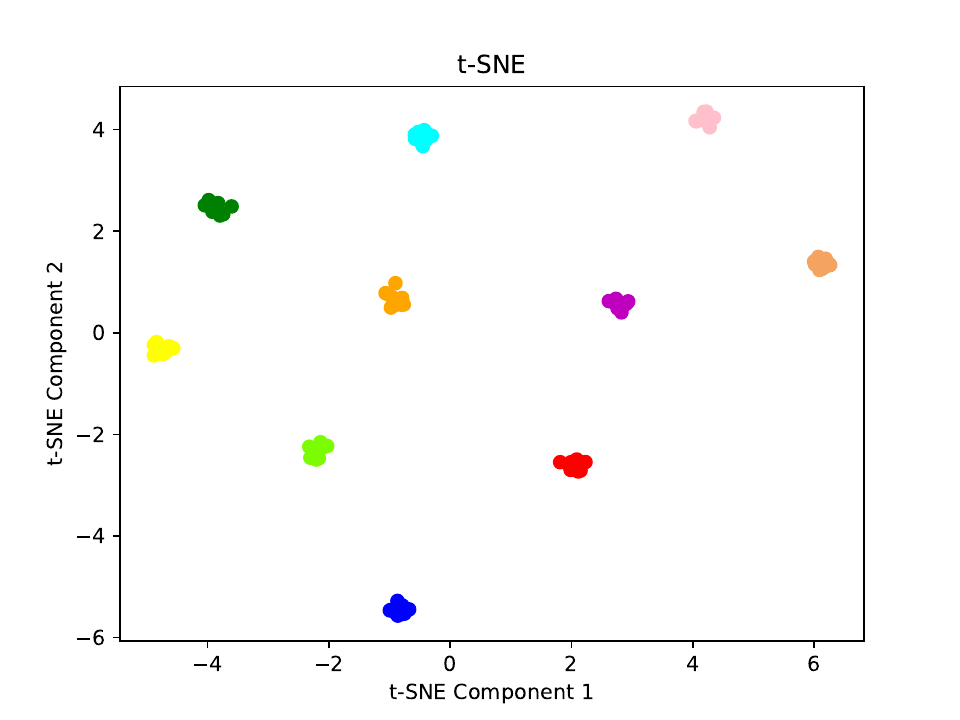}
        \caption{Ours IPC=10}
        \label{fig:STD_10}
      \end{subfigure}
      \begin{subfigure}{0.45\linewidth}
        \includegraphics[width=0.8\linewidth]{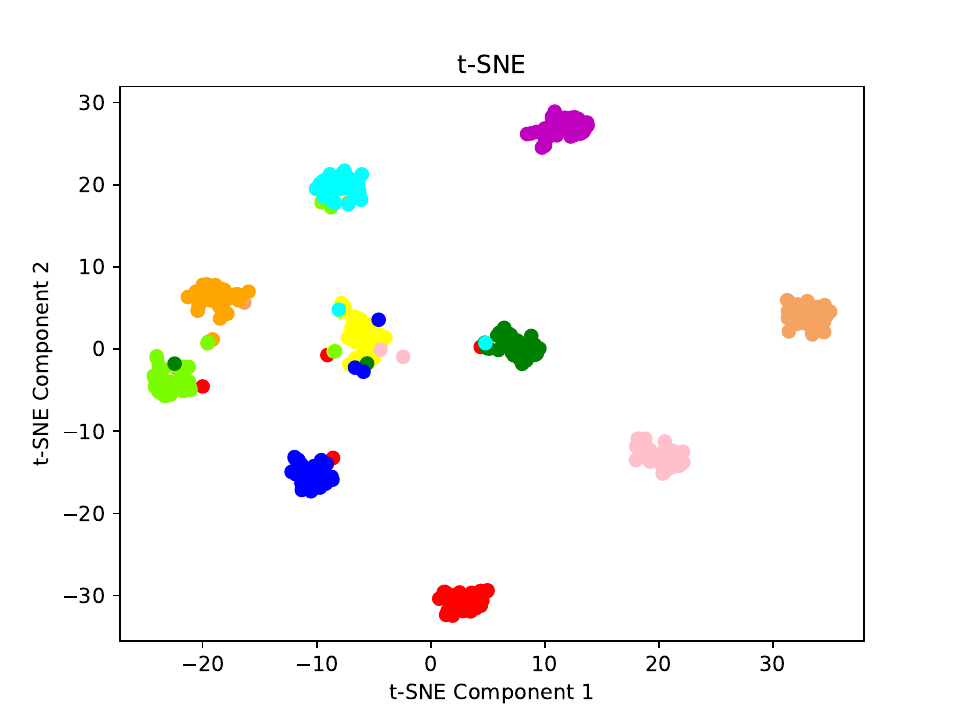}
        \caption{DM IPC=50}
        \label{fig:DM_50}
      \end{subfigure}
      \begin{subfigure}{0.45\linewidth}
        \includegraphics[width=0.8\linewidth]{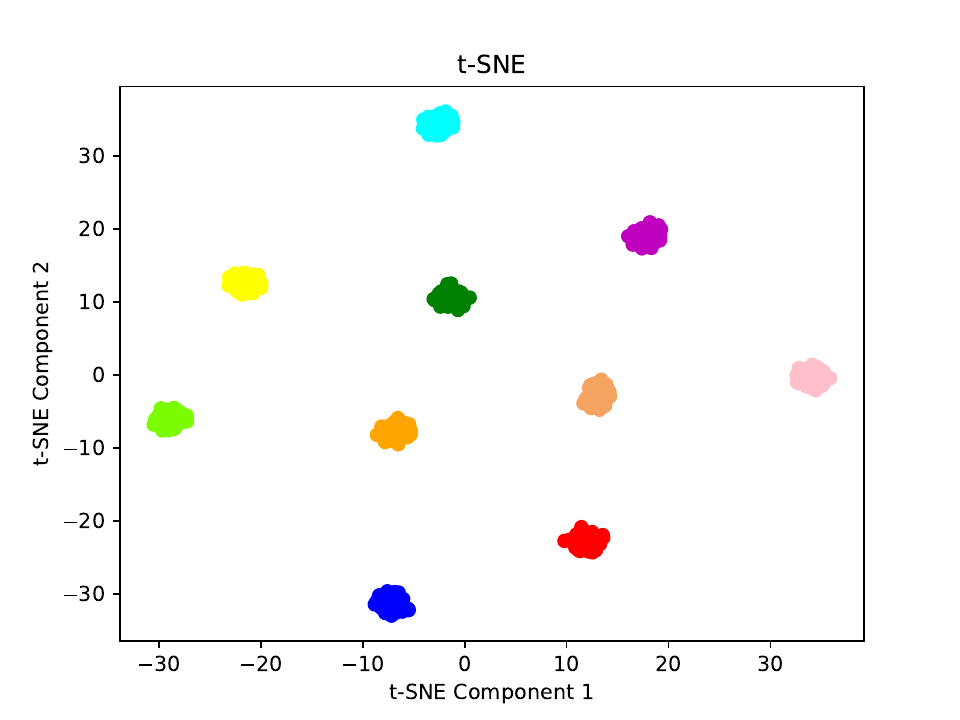}
        \caption{Ours IPC=50}
        \label{fig:STD_50}
      \end{subfigure}
      \caption{T-SNE visualisation of features from synthetic dataset obtained by DM~\cite{zhao2023dataset} and our method, using a pre-trained Resnet18 on CIFAR10. Different colors represent different classes. IPC denotes the number of images per class.} 
      \label{fig:DM_STD_tsne}
    \end{figure}

Dataset distillation, also known as dataset condensation, initially relied on coreset selection methods~\cite{welling2009herding,sener2017active,farahani2009facility}, which involve  selecting representative samples from a dataset. However, these methods have limitations in performance and scalability, especially for large datasets. Pioneering work by Wang~\etal~\cite{wang2018dataset} introduces a meta-learning based approach for dataset distillation, which uses backpropagation to optimize images directly. Presently, dataset distillation methods are primarily categorized into three types: gradient matching~\cite{zhao2020dataset,zhao2021dataset}, trajectory matching~\cite{cazenavette2022dataset}, and distribution matching~\cite{zhao2023dataset,zhao2023improved}. The first two types, despite their effectiveness, are computationally expensive due to their reliance on second-order gradient optimization. In contrast, distribution matching methods~\cite{zhao2023dataset} like those introduced by Zhao~\etal~\cite{zhao2023improved}, CAFE~\cite{wang2022cafe}, and DataDAM~\cite{sajedi2023datadam} address this challenge by matching feature distributions in the embedding space, thereby reducing computational costs. More details about distribution matching method are provided in Section~\ref{sec_3_1}.

In this paper, we focus on distribution matching (DM) methods due to their impressive performance and computational efficiency. The idea of the distribution matching method is to make the synthetic dataset have the same feature distribution as the real dataset with the same class of samples, so how to carve and match this distribution is the key to improve the performance. We identify two key limitations of the methods and design two constraints to improve them. 
First, the feature distribution of samples within the same class in synthetic datasets could still be excessively scattered, leading to poor class discrimination in the embedding space. This issue becomes more pronounced with a smaller number of images per class (IPC), as shown in~\cref{fig:DM_10,fig:DM_50}. To counter this, we propose a \emph{class centralization constraint}, designed to promote clustering of class-specific samples, demonstrated in~\cref{fig:STD_10,fig:STD_50}. 
Second, existing methods inadequately match feature distributions, focusing only on the mean features between real and synthetic datasets. We argue that a comprehensive feature distribution description should include not only means but also covariance matrices, the latter characterises inter-feature relationships. However, in synthetic datasets the number of samples is typically smaller than feature dimensions, accurately estimating covariance matrices is challenging. Our solution is a \emph{covariance matching constraint} that can achieve more precise feature distribution matching between real and synthetic datasets through local feature covariance matrices, even with small sample sizes.

Importantly, both constraints can be seamlessly integrated with existing methods. We have applied them to DM~\cite{zhao2023dataset} and IDM~\cite{zhao2023dataset}, two representative  distribution matching-based methods, and carried out experiments on SVHN, CIFAR10, CIFAR100, and Tiny-Imagenet datasets. Our experiments demonstrate a performance gain of up to 6.6\% on CIFAR10. In cross-architecture experiments, synthetic dataset obtained by the proposed method is tested on network architectures different from the distillation phase, our method exhibits a minimal performance reduction, with a maximum of only 1.7\% across four architectures.

Our main contributions of this work are as follows:
\begin{itemize}
\item We propose a simple yet effective \textit{class centralization constraint}, significantly enhancing class discrimination in synthetic dataset, and improving the performance of dataset distillation.
\item We develop a novel \textit{covariance matching constraint}, which facilitates more accurate  feature distribution matching between real and synthetic datasets even with limited sample sizes, thus effectively augmenting the general mean distribution matching approach.
\item We evaluate the proposed two constraints across multiple benchmark datasets, achieving substantial performance improvements over recent baseline methods and surpassing current state-of-the-art techniques.
\end{itemize}

\section{Related Works}  
{\bf Coreset selection.}
Coreset selection, an early approach to dataset distillation, involves selecting representative samples from a dataset and is applied in various contexts~\cite{aljundi2019gradient,castro2018end,rebuffi2017icarl}. Simple methods include random selection, while more sophisticated techniques like Herding~\cite{welling2009herding} focuses on class centers, and K-Center~\cite{sener2017active,farahani2009facility} selects multiple centroids. The forgetting~\cite{toneva2018empirical} identifies samples that are easily forgotten during training to select representative ones. However, coreset selection struggles with scalability on large datasets and often exhibits suboptimal performance.

{\bf Dataset distillation.}
Dataset distillation or condensation aims to condense a large dataset into much smaller yet informative one. It finds applications in neural architecture search~\cite{du2023minimizing,cui2022dc,zhao2020dataset,zhao2021dataset,zhao2023dataset}, continual learning~\cite{rosasco2021distilled,du2023minimizing,sangermano2022sample}, and privacy protection~\cite{dong2022privacy,zheng2022differentially,carlini2022no}, \etc. The concept was introduced by Wang \etal~\cite{wang2018dataset} using a meta-learning approach. Nguyen \etal~\cite{nguyen2021dataset} further optimize this method using NTK-based ridge regression for better cross-architecture generalization. Zhao~\etal~\cite{zhao2021dataset} further improve the distillation performance using an insertable differentiable siamese augmentation. After that, Bohdal~\etal~\cite{gou2021knowledge} demonstrate that labels could also be distilled. Gradient matching methods proposed by Zhao \etal~\cite{zhao2020dataset} and trajectory matching methods by Cazenavette \etal~\cite{cazenavette2022dataset}  aim to match gradients and model parameters of synthetic and real datasets, respectively. Recent advances~\cite{zhao2022synthesizing,deng2022remember,lee2022dataset} focus on preserving features and corresponding decoders, then restore the features to the original size of the image. This will enhance performance while also introducing additional preprocessing steps for synthetic datasets.

To reduce the high computational cost of the previous methods, Zhao~\etal~\cite{zhao2023dataset} propose distribution matching (DM), optimizing for maximum mean discrepancy (MMD) between synthetic and real datasets.  CAFE~\cite{wang2022cafe} aligns features by matching them across layers in convolutional neural networks, and IDM~\cite{zhao2023improved} improves DM with partitioning augmentation and class-aware distribution regularization. DataDAM~\cite{sajedi2023datadam} proposes an attention matching framework to dataset distillation by focusing on feature attention. 

Although demonstrating promising performance, existing distribution matching methods methods face two primary limitations: the dispersed feature distribution within the same class in synthetic datasets, reducing class discrimination, and an exclusive focus on mean feature consistency, lacking precision and comprehensiveness. This work addresses above limitations by proposing two plug-and-play constraints, which significantly enhance dataset distillation performance by focusing on inter-sample and inter-feature relations, respectively.
\section{Method}

In this section, we outline the fundamentals of dataset distillation and basic framework of distribution matching-based methods. We then delve into specifics of our proposed \emph{class centralization constraint} and \emph{covariance matching constraint}. Lastly, we present the entire objective function.

\subsection{Preliminaries}
\label{sec_3_1}
Given a large image dataset $\mathcal{T}=\{(x_1,y_1), (x_2,y_2), \ldots, \\
(x_{\vert \mathcal{T}\vert},y_{\vert \mathcal{T}\vert})\}$, containing $\vert \mathcal{T}\vert$ images and labels, dataset distillation aims to condense $\mathcal{T}$ to a significantly smaller synthetic dataset $\mathcal{S}=\{(s_1,y_1^{s}), (s_2,y_2^{s}), \ldots, (s_{\vert \mathcal{S}\vert},y_{\vert \mathcal{S}\vert}^{s})\}$ comprising $\vert \mathcal{S}\vert$ images and class labels. The goal is to ensure that a model $\psi_{{\theta}^\mathcal{S}}$, trained from scratch with $\mathcal{S}$, performs comparably to a model $\psi_{{\theta}^\mathcal{T}}$ trained with $\mathcal{T}$. This is achieved by optimizing the following objective function:
    \begin{equation}\small
    \mathcal{S}^*=\underset{\mathcal{S}}{\arg \min } \mathbb{E}_{\boldsymbol{x} \sim P_{\mathcal{T}}}\left\|\ell\left(\psi_{\boldsymbol{\theta}^{\mathcal{T}}}(\boldsymbol{x}), y\right)-\ell\left(\psi_{\boldsymbol{\theta}^{\mathcal{S}}}(\boldsymbol{x}), y\right)\right\|\,,
    \end{equation}
where $\ell$ represents the cross-entropy loss since all existing dataset distillation methods focus on classification tasks.

DM~\cite{zhao2023dataset} is a representative distribution matching-based dataset distillation method, aiming to minimize the maximum mean discrepancy (MMD) between the synthetic dataset and real dataset:
    \begin{equation}\small
    \label{eq:dm}
    \mathbb{E}_{\boldsymbol{\theta} \sim P_{\boldsymbol{\theta}}}\left\|\frac{1}{|\mathcal{T}|} \sum_{i=1}^{|\mathcal{T}|} \psi_{\boldsymbol{\theta}}\left(\boldsymbol{x}_i\right)-\frac{1}{|\mathcal{S}|} \sum_{j=1}^{|\mathcal{S}|} \psi_{\boldsymbol{\theta}}\left(\boldsymbol{s}_j\right)\right\|^2\,,
    \end{equation}
where $P_{\boldsymbol{\theta}}$ denotes the distribution of randomly initialized network parameters.

\subsection{Class centralization constraint (inter-sample)}
Synthetic datasets obtained by methods like DM~\cite{zhao2023dataset} often exhibit insufficient class discrimination. This issue, particularly evident when the number of samples per class (IPC) is small, is illustrated by the scattered feature distribution and unclear class boundaries in synthetic datasets (\cref{fig:DM_10,,fig:DM_50}). To address this, we propose a class centralization constraint, which aims to cluster features $\phi (s)$ extracted from the synthetic dataset within the same class, rather than allowing them to disperse. We formulate the loss to enforce this constraint as:
    \begin{equation}\small
    \label{eq:cluster}
        \mathcal{L}_{CC} =  \sum \limits _{c} ^{C}(\sum \limits _{j=1} ^{K}  \max(0,\exp(\alpha \left\| \psi (s^c_j)-\bar{\psi(s^c)} \right\|^2)-\beta)),
    \end{equation}
    where
    \begin{equation}\small
        \bar{\psi(s^c)} = \frac{1}{K} \sum \limits _{j=1} ^{K} \psi (s^c_j),
    \end{equation}
$C$ is the number of categories, $K$ is the number of samples per class, $\alpha$ is a scaling factor, and $\beta$ is the centralization threshold. A smaller $\beta$ encourages tighter clustering of samples within each class.

Since this is a plug-and-play constraint, we keep the original constraint~\cref{eq:dm} from the baseline, but it's worth noting that the previous methods used ConvNet $\psi$ in~\cref{eq:dm} with randomly initialized parameters (DM~\cite{zhao2023dataset}) or pre-trained parameters (IDM~\cite{zhao2023improved}), our approach employs Resnet18 for network $\psi$ in~\cref{eq:cluster}, which is distinct from network $\psi$ in~\cref{eq:dm}. Despite ResNet's more complex architecture, which typically poses challenges for small datasets, our experiments reveal that Resnet18 more effectively differentiates between class features. Additionally, using different neural networks enhances the cross-architecture generalization of dataset distillation. As illustrated in~\cref{fig:STD_10,fig:STD_50}, our class centralization constraint results in a more concentrated feature distribution within each class in the synthetic dataset, achieving clear differentiation between classes. The effectiveness of this constraint is further validated in our experimental results.

    \begin{figure}
      \centering
      \includegraphics[width=1.0\linewidth]{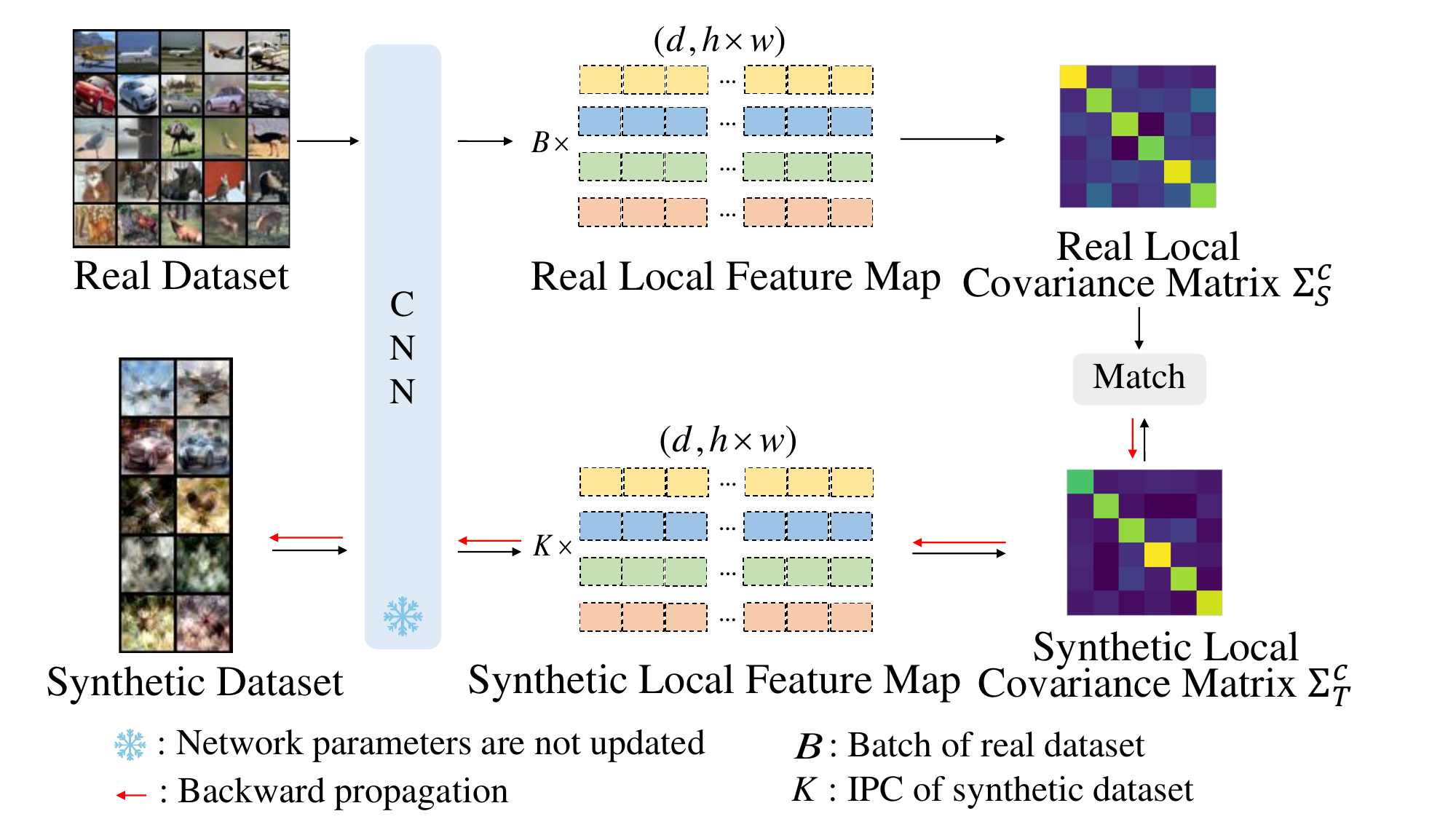}
      \caption{Illustration of the proposed covariance matching constraint. This constraint involves calculating local covariance matrices for corresponding classes in both real and synthetic datasets, followed by matching these matrices.}
      \label{fig:localfig}
    \end{figure}

\subsection{Covariance matching constraint (inter-feature)}
The essence of distribution matching-based dataset distillation lies in aligning the feature distributions of real and synthetic datasets. While existing methods primarily focus on matching feature means, an effective representation of feature distribution shall also consider the covariance matrix, which captures inter-feature relationships. However, in distilled synthetic datasets, the number of samples in each class is often much smaller than the feature dimensions. This is often known as ``small sample problem" and it could lead to less precise covariance matrix estimations.

To address this, we propose the covariance matching constraint for more accurate feature distribution matching, even when the sample size is smaller than the dimensions of features extracted by network $\psi$. As shown in~\cref{fig:localfig}, for the feature of a single sample, rather than flattening, we reshape it into a tensor of shape $(d,hw)$, resulting in $d$ $hw$-dimensional local feature descriptors, expressed as $X_i\in \mathbb{R}^{d\times hw}$ for real dataset,  $S_i\in \mathbb{R}^{d\times hw}$ for synthetic dataset. In this way, we greatly reduce the feature dimensions and avoid high-dimensional vector space computation, which facilitates a more accurate evaluation of the covariance matrices.
We then calculate the local feature covariance matrices $\Sigma_\tau\ \in \mathbb{R}^{d\times d}$ and $\Sigma_s \in \mathbb{R}^{d\times d}$ for the real and synthetic datasets, respectively, and compute a matching loss between these two matrices:
    \begin{equation}\small
    \label{eq:local}
    \mathcal{L}_{CM} =   \sum \limits _{c=1} ^{C} \left\| \Sigma^c_s-\Sigma^c_\tau \right\|^2,
    \end{equation}
    where
    \begin{equation}\small
    \Sigma^c_s =\frac{1}{K} \sum \limits _{i=1} ^{K} (S^c_i-\bar{S^c}) (S^c_i-\bar{S^c})^\top,  
    \end{equation}
    \begin{equation}\small
    \Sigma^c_\tau =\frac{1}{B} \sum \limits _{i=1} ^{B} (X^c_i-\bar{X^c}) (X^c_i-\bar{X^c})^\top.      
    \end{equation}
Here, $K$ is the number of samples per class in the synthetic dataset $\mathcal{S}$, $B$ is the batch size per class of the real dataset $\mathcal{T}$, $\bar{S^c}\in\mathbb{R}^{d\times hw}$ and $\bar{X^c}\in\mathbb{R}^{d\times hw} $ are the feature means of class $c$ for the synthetic and real datasets, respectively.

\subsection{The objective function}
Our proposed constraints are designed as plug-and-play, making them adaptable to various distribution matching-based methods. For instance, when employing DM~\cite{zhao2023dataset} as the baseline method, the overall objective function can be expressed as:
    \begin{equation}\small
    \label{eq:overalldm}
    \mathcal{L} = \mathcal{L}_{DM}+\lambda_{CC}\mathcal{L}_{CC}+\lambda_{CM}\mathcal{L}_{CM}\,.
    \end{equation}
Similarly, when using IDM~\cite{zhao2023improved} as the baseline, the objective function will be defined as:
    \begin{equation}\small
    \label{eq:overallidm}
    \mathcal{L} = \mathcal{L}_{IDM}+\lambda_{CC}\mathcal{L}_{CC}+\lambda_{CM}\mathcal{L}_{CM}\,.
    \end{equation}
In above formulas, $\lambda_{CC}$ and $\lambda_{CM}$ are the weighting parameters, and $\mathcal{L}_{DM}$ and $\mathcal{L}_{IDM}$ represent the loss functions in DM~\cite{zhao2023dataset} and IDM~\cite{zhao2023improved}, respectively.
\section{Experiments}

    \begin{table*}[t] 
    \small
    \centering
    \tabcolsep=4pt
    \caption{Comparative analysis of dataset distillation methods. Ratio (\%): the proportion of condensed images relative to the number of entire training set. Whole Dataset: the accuracy of training on the entire original dataset. The \textbf{best} results in each column are highlighted for clarity.}
    \label{tab:main}
    \begin{tabular}{l *{10}{c}}
        \toprule
        Method &  Venue & \multicolumn{3}{c}{SVHN} & \multicolumn{3}{c}{CIFAR10} & \multicolumn{3}{c}{CIFAR100}  \\
            \midrule
        IPC &  & 1 & 10 & 50 & 1 & 10 & 50 & 1 & 10 & 50 \\
        Ratio (\%) &  & 0.0014 & 0.14 & 0.7 & 0.02 & 0.2 & 1 & 0.2 & 2 & 10 \\
        \midrule
        Random & Classic & 14.6$\pm$1.6 & 35.1$\pm$4.1 & 70.9$\pm$0.9 & 14.4$\pm$2.0 & 26.0$\pm$1.2 & 43.4$\pm$1.0 & 4.2$\pm$0.3 & 14.6$\pm$0.5 & 30.0$\pm$0.4 \\
        Herding~\cite{welling2009herding} & Classic & 20.9$\pm$1.3 & 50.5$\pm$3.3 & 72.6$\pm$0.8 & 21.5$\pm$1.2 & 31.6$\pm$0.7 & 40.4$\pm$0.6 & 8.4$\pm$0.3 & 17.3$\pm$0.3 & 33.7$\pm$0.5 \\
        K-Center~\cite{sener2017active} & Classic & 21.0$\pm$1.5 & 14.0$\pm$1.3 & 20.1$\pm$1.4 & 21.5$\pm$1.3 & 14.7$\pm$0.9 & 27.0$\pm$1.4 & 8.4$\pm$0.3 & 17.3$\pm$0.2 & 30.5$\pm$0.3 \\
        Forgetting~\cite{toneva2018empirical} & Classic & 12.1$\pm$5.6 & 16.8$\pm$1.2 & 27.2$\pm$1.5 & 13.5$\pm$1.2 & 23.3$\pm$1.0 & 23.3$\pm$1.1 & 4.5$\pm$0.3 & 15.1$\pm$0.2 & 30.5$\pm$0.4 \\
        \midrule
        DD~\cite{wang2018dataset} & Arxiv'18 & - & - & - & - & 36.8$\pm$1.2 & - & - & - & -\\
        LD~\cite{bohdal2020flexible} & NeurIPS'20 & - & - & - & 25.7$\pm$0.7 & 38.3$\pm$0.4 & 42.5$\pm$0.4 & 11.5$\pm$0.4 & - & - \\
        DC~\cite{zhao2020dataset} & ICLR'21  & 31.2$\pm$1.4 & 76.1$\pm$0.6 & 82.3$\pm$0.3 & 28.3$\pm$0.5 & 44.9$\pm$0.5 & 53.9$\pm$0.5 & 12.8$\pm$0.3 & 25.2$\pm$0.3 & - \\
        DSA~\cite{zhao2021dataset} & ICML'21 & 27.5$\pm$1.4 & 79.2$\pm$0.5 & 84.4$\pm$0.4 & 28.8$\pm$0.7 & 52.1$\pm$0.5 & 60.6$\pm$0.5 & 13.9$\pm$0.3 & 32.3$\pm$0.3 & 42.8$\pm$0.4 \\
        KIP~\cite{nguyen2021dataset} & NeurIPS’21 & 62.4$\pm$0.3 & 81.1$\pm$0.5 & 84.3$\pm$0.1 & 29.8$\pm$0.2 & 46.1$\pm$0.3 & 53.2$\pm$0.4 & 12.0$\pm$0.2 & 29.0$\pm$0.2 & - \\
        CAFE~\cite{wang2022cafe} & CVPR'22 & 42.9$\pm$3.3 & 77.9$\pm$0.6 & 82.3$\pm$0.3 & 30.3$\pm$1.1 & 46.3$\pm$0.6 & 55.5$\pm$0.6 & 12.9$\pm$0.3 & 27.8$\pm$0.3 & 37.9$\pm$0.3 \\
        DCC~\cite{lee2022datasetsignal} & ICML'22 & 34.3$\pm$1.6 & 76.2$\pm$0.8 & 83.3$\pm$0.2 & 34.0$\pm$0.7 & 54.4$\pm$0.5 & 64.2$\pm$0.4 & 14.6$\pm$0.3 & 33.5$\pm$0.3 & 39.3$\pm$0.4 \\
        DataDAM~\cite{sajedi2023datadam} & ICCV'23 & - & - & - & 32.0$\pm$1.2 & 54.2$\pm$0.8 & 67.0$\pm$0.4 & 14.5$\pm$0.5 & 34.8$\pm$0.5 & 49.4$\pm$0.3 \\
        \midrule
        DM~\cite{zhao2023dataset} & WACV'23 & 21.6$\pm$0.8 & 72.8$\pm$0.3 & 82.6$\pm$0.5 & 26.4$\pm$0.8 & 48.5$\pm$0.6 & 62.2$\pm$0.5 & 11.4$\pm$0.3 & 29.7$\pm$0.3 & 43.0$\pm$0.4 \\
        \textbf{DM+Ours} &-  & - & 75.7$\pm$0.3 & \textbf{85.3$\pm$0.2} & - & 55.1$\pm$0.1 & 65.1$\pm$0.2 & - & 32.2$\pm$0.5 & 43.6$\pm$0.3   \\
        \midrule
         IDM~\cite{zhao2023improved} & CVPR'23 & 65.3$\pm$0.3 & 81.0$\pm$0.1 & 84.1$\pm$0.1 & 45.2$\pm$0.5 & 57.3$\pm$0.3 & 67.2$\pm$0.1 & 23.1$\pm$0.2 & 44.7$\pm$0.1 & 49.9$\pm$0.2 \\
        \textbf{IDM+Ours} &-  & \textbf{66.3$\pm$0.1} & \textbf{82.1$\pm$0.3} & 85.1$\pm$0.5 & \textbf{47.1$\pm$0.1} & \textbf{59.9$\pm$0.2} & \textbf{69.0$\pm$0.3} & \textbf{24.6$\pm$0.1} & \textbf{45.7$\pm$0.4} & \textbf{51.3$\pm$0.4} \\
        \midrule
        Whole Dataset &   & \multicolumn{3}{c}{95.4$\pm$0.2} & \multicolumn{3}{c}{84.8$\pm$0.1} & \multicolumn{3}{c}{56.2$\pm$0.3}  \\
        \bottomrule
    \end{tabular}
    \end{table*}

\subsection{Experimental setup}

\textbf{Datasets.} We evaluate our method on several benchmark datasets for dataset distillation: SVHN~\cite{sermanet2012convolutional}, CIFAR10 and CIFAR100~\cite{krizhevsky2009learning}, as well as the larger TinyImageNet~\cite{le2015tiny}. SVHN contains over 600,000 digit images of house numbers around the world. CIFAR10 and CIFAR100 feature 10 and 100 classes respectively, each with 600 images per class, and an image resolution of $32\times 32$. For larger datasets, we performed on TinyImageNet~\cite{le2015tiny}, which contains 200 classes with 600 images in each class, which have an image resolution of 64×64.


\textbf{Network architectures.} For $\mathcal{L}_{DM}$, $\mathcal{L}_{IDM}$ in \cref{eq:overalldm} and \cref{eq:overallidm}, we utilize the ConvNet with the same architecture as in DM~\cite{zhao2023dataset} and IDM~\cite{zhao2023improved} to extract features. This architecture comprises 3 blocks, each with a $3\times 3$ convolutional layer, instance normalization, ReLU activation, and a $3\times 3$ average pooling with a stride of 2. For our class centralization constraint  $\mathcal{L}_{CC}$, we employ a pre-trained ResNet18~\cite{he2016deep}, trained for 30 epochs on each corresponding dataset. The architecture of ResNet18 includes 4 layers with 2 blocks per layer, and convolutional layer, instance normalization, and ReLU activation. ConvNet, AlexNet, VGG11, and ResNet18 
are used in the cross-architecture generalization experiments, and we follow the architectural setup of DM~\cite{zhao2023dataset}.


\textbf{Evaluation.} We adhere to the evaluation protocol of previous works~\cite{lee2022dataset,wang2018dataset,zhao2021dataset,zhao2023dataset,zhao2020dataset}. On various datasets, we synthesize distilled datasets with IPC = $1,10,50$, then train a randomly initialized model from scratch using these datasets. The training configurations follow the prior works~\cite{lee2022dataset,wang2018dataset,zhao2021dataset,zhao2023dataset,zhao2020dataset}. We compute the Top-1 accuracy on the test set of the original real dataset, repeating each experiment five times to calculate the mean value.

\textbf{Implementation details.} Our method is implemented based on DM~\cite{zhao2023improved} and IDM~\cite{zhao2023improved}, adhering to their baseline hyperparameter settings. We follow the DSA~\cite{zhao2021dataset} augmentation method used in previous works. Synthetic datasets are learned using SGD  with a learning rate of 1, and a batch size of 5000 for IDM experiments with IPC=1 on CIFAR10/100, and 256 in other scenarios. We set $\lambda_{CC}$ as 0.05, and $\lambda_{CM}$ as 0.01 for IPC = 1, and 10, while for IPC = 50, $\lambda_{CC}$ is set as 0.003 and $\lambda_{CM}$ as 0.01. All experiments are conducted using the PyTorch framework on two RTX3090 GPUs, each with 24 GB of memory.

\subsection{Comparison with state-of-the-art methods}

\textbf{Competitive methods.} We compare our method with both classic coreset selection methods, including Random, Herding~\cite{welling2009herding}, K-Center~\cite{sener2017active}, and Forgetting~\cite{toneva2018empirical}, and state-of-the-art dataset distillation techniques including DD~\cite{wang2018dataset}, LD~\cite{bohdal2020flexible}, DC~\cite{zhao2020dataset}, DSA~\cite{zhao2021dataset}, KIP~\cite{nguyen2021dataset}, CAFE~\cite{wang2022cafe}, DCC~\cite{lee2022datasetsignal}, DM~\cite{zhao2023dataset}, IDM~\cite{zhao2023improved}, DataDAM~\cite{sajedi2023datadam}. For DM~\cite{zhao2023dataset} and IDM~\cite{zhao2023improved}, we report results reproduced using their official codes, while results for other methods are quoted from their respective original papers.

\textbf{Performance comparison.} 
\cref{tab:main,tab:maintiny} show the results of our method in comparison with previous methods across various datasets.
The experimental results demonstrate significant improvements over the baselines. Specifically, our method surpasses DM by 2.9\% and 2.7\% on SVHN at IPC=10 and 50, respectively;  6.6\% and 2.9\%  on CIFAR10 at IPC=10 and 50; and 2.5\% and 2.0\% on TinyImageNet at IPC=10 and 50. When compared with IDM, our method shows an improvement of  1.9\%, 2.6\% and 1.8\% on CIFAR10 at IPC=1, 10, and 50, respectively, and 1.5\%, 1.0\%, and 1.4\% on CIFAR100 at IPC=10, 50. Our IDM-based method also outperforms all other state-of-the-art methods, achieving accuracies of 59.9\% and 69.0\% on CIFAR10, and 45.7\% and 51.3\% on CIFAR100, both at IPC=10 and 50. On the larger TinyImageNet dataset, our method also achieves state-of-the-art performance with an accuracy of 23.3\% at IPC=10, surpassing  the second best method by 1.3\%. For the simpler SVHN dataset with IPC=10, using DM as the baseline proved most effective, suggesting that simpler approaches may be more beneficial for less complex datasets. 

 Notably, our method, when used with DM as the baseline, does not provide IPC=1 results due to its requirement for a sample size greater than 1. However, IDM, with its ability to preserve more samples at IPC=1 through partitioning and expansion augmentation, is compatible with our approach.

    \begin{table}[t]\small
    \centering
    \centering
    \tabcolsep=4pt
    \caption{Comparative analysis of dataset distillation methods on higher resolution datasets. The \textbf{best} results in each column are highlighted for clarity.}
    \label{tab:maintiny}
    \begin{tabular}{l *{4}{c}}
        \toprule
        Method &  Venue  & \multicolumn{3}{c}{TinyImageNet} \\
            \midrule
        IPC &  &  1 & 10 & 50\\
        Ratio (\%) &  & 0.2 & 2 & 10\\
        \midrule
        Random & Classic &  1.4$\pm$0.1 & 5.0$\pm$0.2 & 15.0$\pm$0.4\\
        Herding~\cite{welling2009herding} & Classic & 2.8$\pm$0.2 & 6.3$\pm$0.2 & 16.7$\pm$0.3\\
        K-Center~\cite{sener2017active} & Classic & 1.6$\pm$0.2 & 5.1$\pm$0.1 & 15.0$\pm$0.3\\
        Forgetting~\cite{toneva2018empirical} & Classic & 1.6$\pm$0.2 & 5.1$\pm$0.3 & 15.0$\pm$0.1\\
        DataDAM~\cite{sajedi2023datadam} & ICCV'23 & 8.3$\pm$0.4 & 18.7$\pm$0.3 & \textbf{28.7$\pm$0.3}\\
        \midrule
        DM~\cite{zhao2023dataset} & WACV'23 & 3.9$\pm$0.2 & 12.9$\pm$0.4 & 24.1$\pm$0.3\\
        \textbf{DM+Ours} & - & - & 15.4$\pm$0.2 & 26.1$\pm$0.1  \\
        \midrule
         IDM~\cite{zhao2023improved} & CVPR'23 & 9.8$\pm$0.2 & 21.9$\pm$0.2 & 26.2$\pm$0.3\\
        \textbf{IDM+Ours} & - &  \textbf{10.0$\pm$0.1} & \textbf{23.3$\pm$0.1} & 27.5$\pm$0.3 \\
        \midrule
        Whole Dataset &   & \multicolumn{3}{c}{37.6$\pm$0.6} \\
        \bottomrule
    \end{tabular}
    \end{table}

\subsection{Cross-architecture generalization}
Cross-architecture generalization is a crucial metric for evaluating dataset distillation, particularly because it is hard to predict the neural network architectures that will be used in a real application. Significant performance drop upon changing model architectures is undesirable. We evaluate the synthetic dataset generated by our method on CIFAR10 across four different architectures, comparing it with previous methods. \cref{tab:cross10,tab:cross50} present the cross-architecture performance for IPC=10 and 50. Our experiments included widely-used models such as ConvNet, AlexNet~\cite{deng2009imagenet}, VGG11~\cite{simonyan2014very}, and ResNet18~\cite{he2016deep}, with architectural details following previous work. Each experiment is repeated five times to determine the mean value.


For IPC=50 as shown in \cref{tab:cross50}, our method's performance on ResNet18 shows an 8\% improvement over DataDAM~\cite{sajedi2023datadam}. More notably, across  ConvNet, AlexNet, VGG11, and ResNet18, we record accuracies of 69.0\%, 67.3\%, 67.3\%, and 68.3\%, respectively, maintaining a performance drop of less than 1.7\% across these architectures. This is particularly significant considering that previous methods often experienced substantial performance drops on ResNet18. For IPC=10 as shown in \cref{tab:cross10}, our method performs significantly better than others, with a performance loss not exceeding 3.1\% on AlexNet, VGG11, and ResNet18, while KIP~\cite{nguyen2021dataset} loss 10.8\% from ConvNet to Resnet18. The better cross-architecture generalization performance achieved by our method allows us to utilize more diverse model architectures in the dataset distillation phase perhaps facilitating the synthesis of datasets that are more generalizable across different model architectures.

Note that some existing methods only appear in one of the two tables because they only reported the result for either IPC = 10 or IPC = 50.
    \begin{table}[!tp]\small
    \centering
    \tabcolsep=4pt
    \caption{Cross-architecture testing on CIFAR10 with IPC=10.}
    \label{tab:cross10}
    \begin{tabular}{l*{5}{c}}
        \toprule
                 & ConvNet & AlexNet & VGG11 & ResNet18 \\
        \midrule
        DM~\cite{zhao2023dataset} &  48.9$\pm$0.6 & 38.8$\pm$0.5 & 42.1$\pm$0.4 & 41.2$\pm$1.1 \\
        DSA~\cite{zhao2021dataset} & 52.1$\pm$0.7 & 35.9$\pm$2.6 & 43.2$\pm$0.5 & 42.8$\pm$1.0 \\
        KIP~\cite{nguyen2021dataset} & 47.6$\pm$0.9 & 24.4$\pm$3.9 & 42.1$\pm$0.4 & 36.8$\pm$1.0 \\
        IDM~\cite{zhao2023improved} & 53.0$\pm$0.3 & 44.6$\pm$0.8 & 47.8$\pm$1.1 & 44.6$\pm$0.4 \\
        \midrule
        \textbf{IDM+Ours}  & \textbf{59.9.$\pm$0.2} & \textbf{56.8$\pm$0.02} & \textbf{58.0$\pm$0.5} & \textbf{56.9$\pm$0.5} \\
        \bottomrule
    \end{tabular}
    \end{table}

\subsection{Ablation study}
\label{sec:ablation exp}
\begin{figure*}[htp]
      \centering
        \begin{subfigure}{0.18\linewidth}
            \includegraphics[width=1\linewidth]{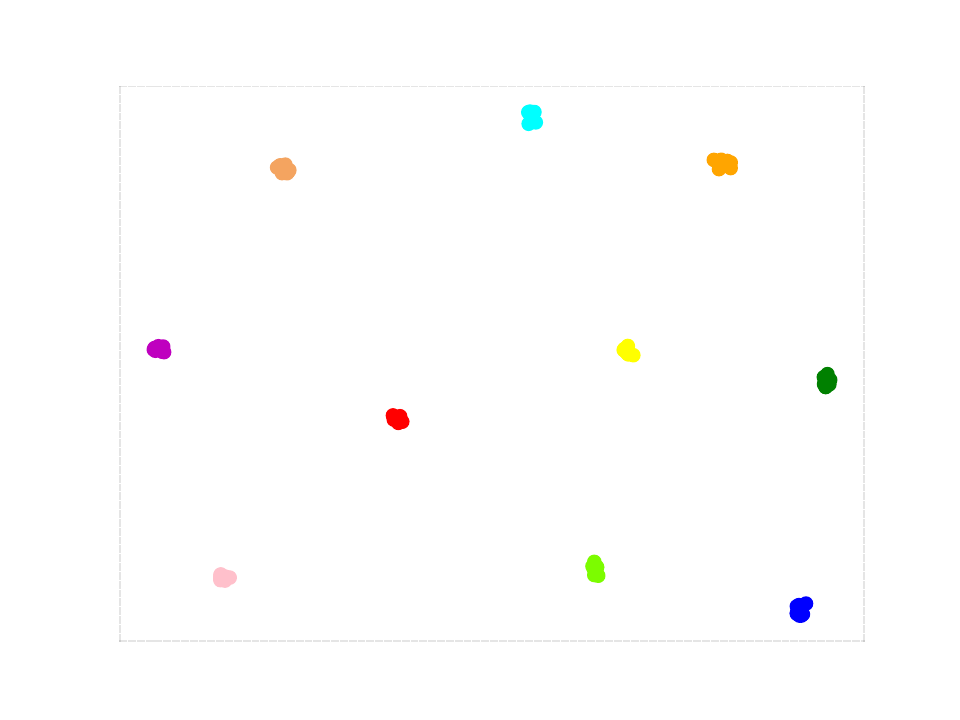}
            \caption{$\beta=0.0$}
        \end{subfigure}
        \hfill
        \begin{subfigure}{0.18\linewidth}
            \includegraphics[width=1\linewidth]{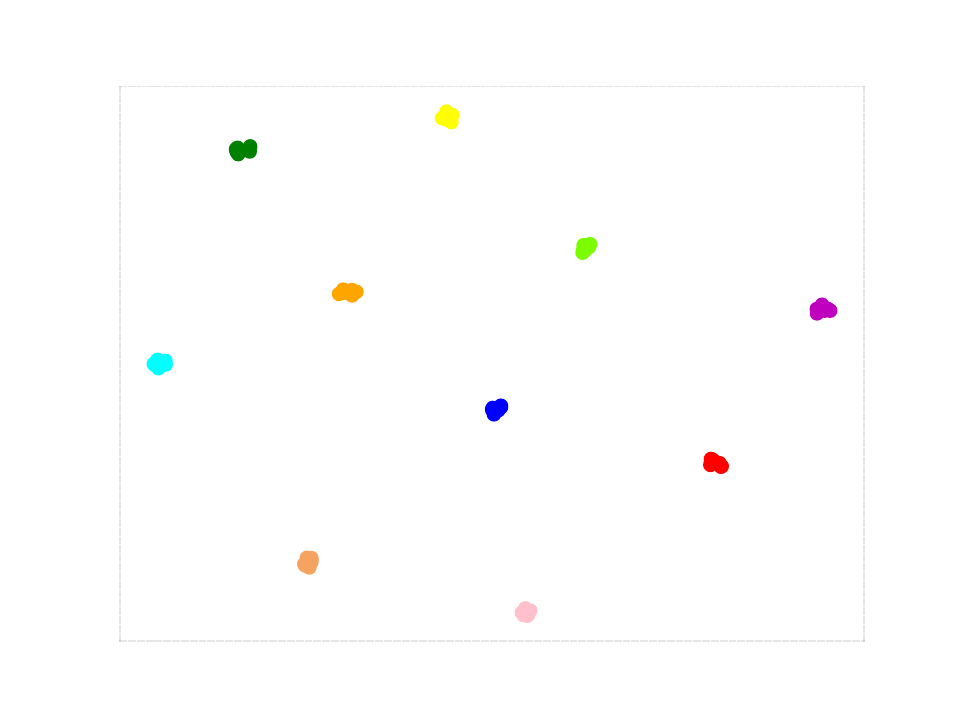}
            \caption{$\beta=0.1$}
        \end{subfigure}
        \hfill
        \begin{subfigure}{0.18\linewidth}
            \includegraphics[width=1\linewidth]{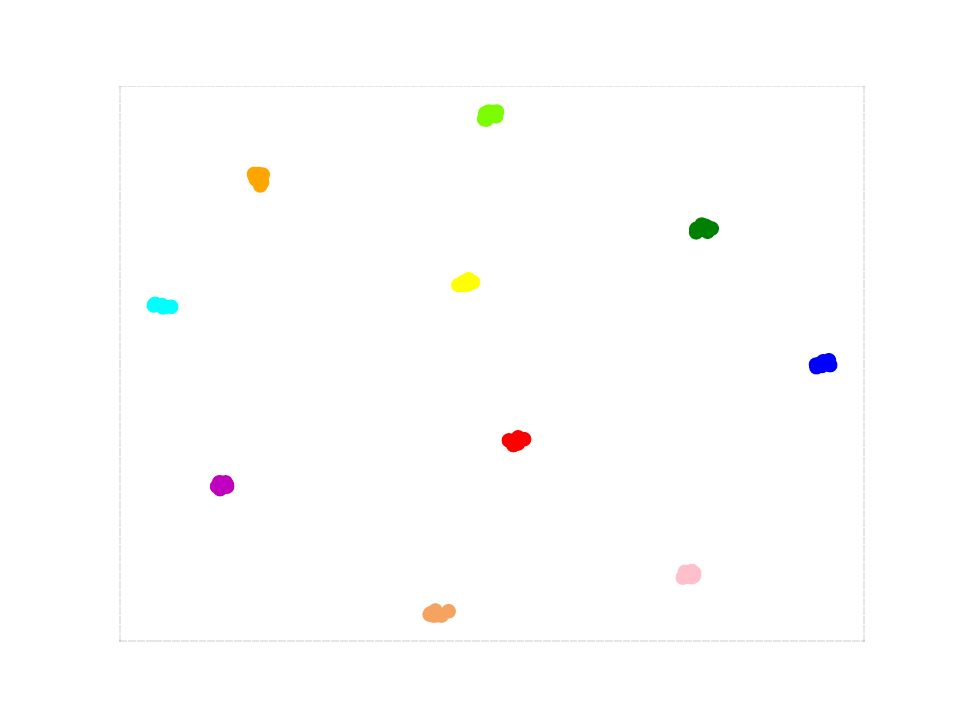}
            \caption{$\beta=0.5$}
        \end{subfigure}
        \hfill
        \begin{subfigure}{0.18\linewidth}
            \includegraphics[width=1\linewidth]{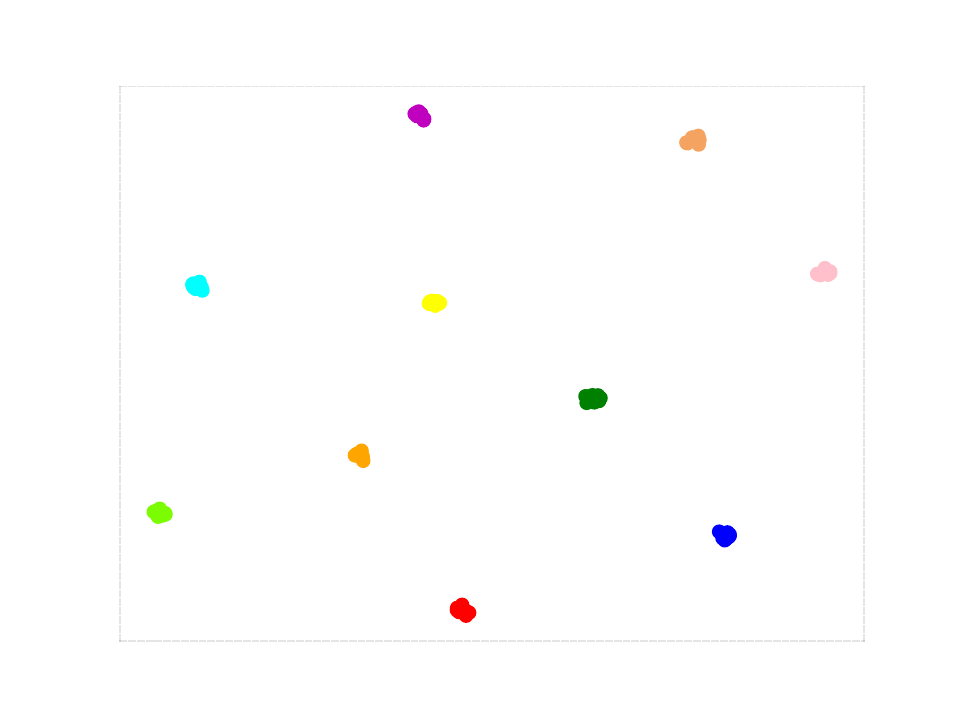}
            \caption{$\beta=0.7$}
        \end{subfigure}
        \hfill
        \begin{subfigure}{0.18\linewidth}
            \includegraphics[width=1\linewidth]{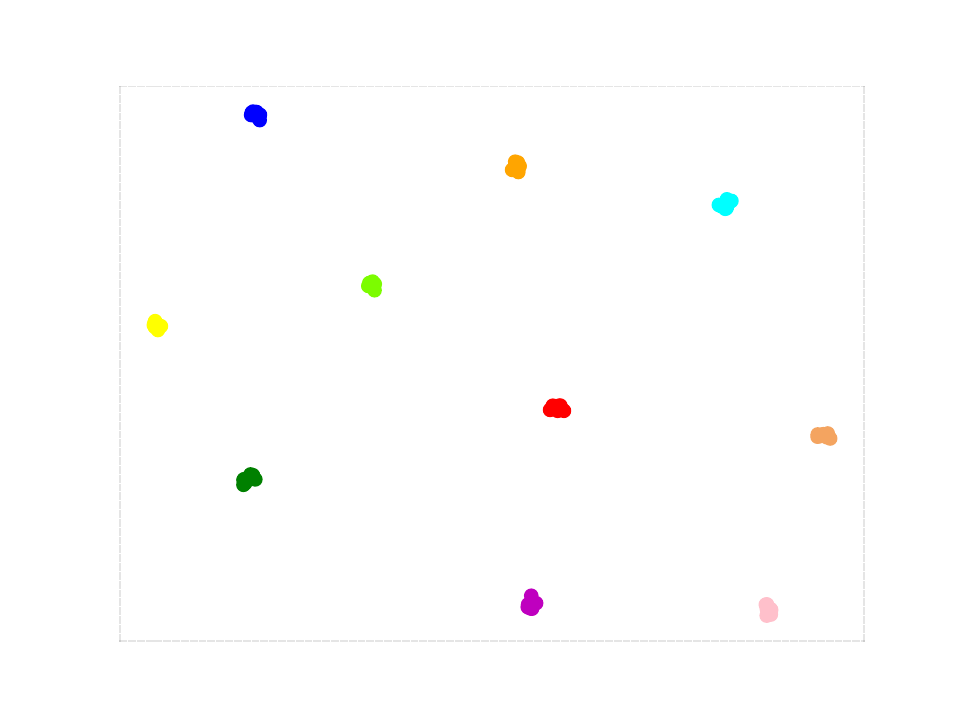}
            \caption{$\beta=0.9$}
        \end{subfigure}
        \hfill
        \begin{subfigure}{0.18\linewidth}
            \includegraphics[width=1\linewidth]{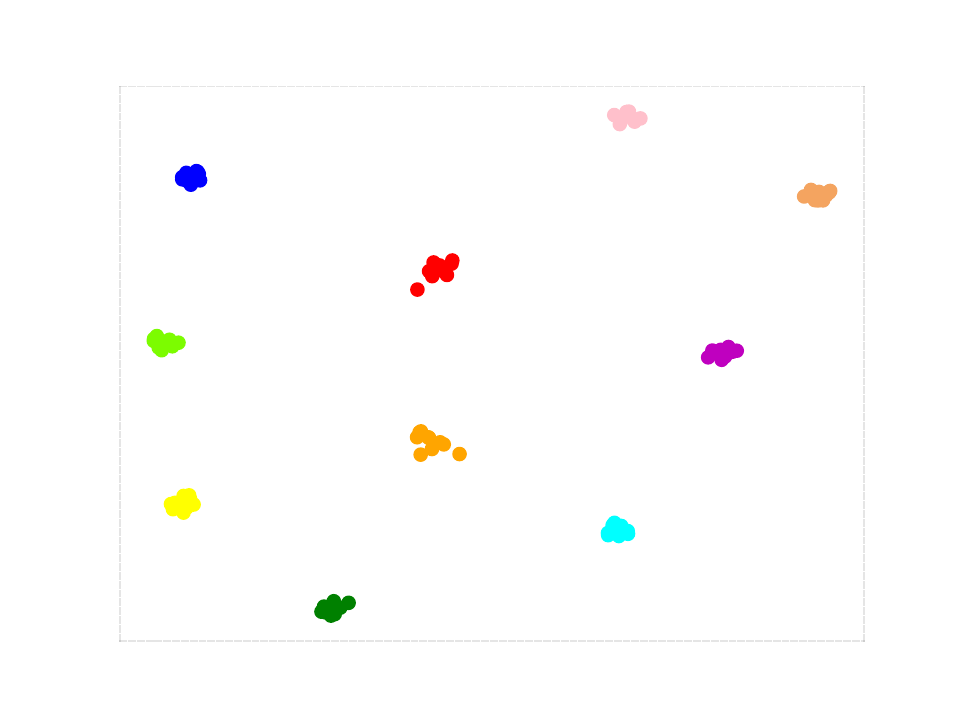}
            \caption{$\beta=1.1$}
        \end{subfigure}
        \hfill
        \begin{subfigure}{0.18\linewidth}
            \includegraphics[width=1\linewidth]{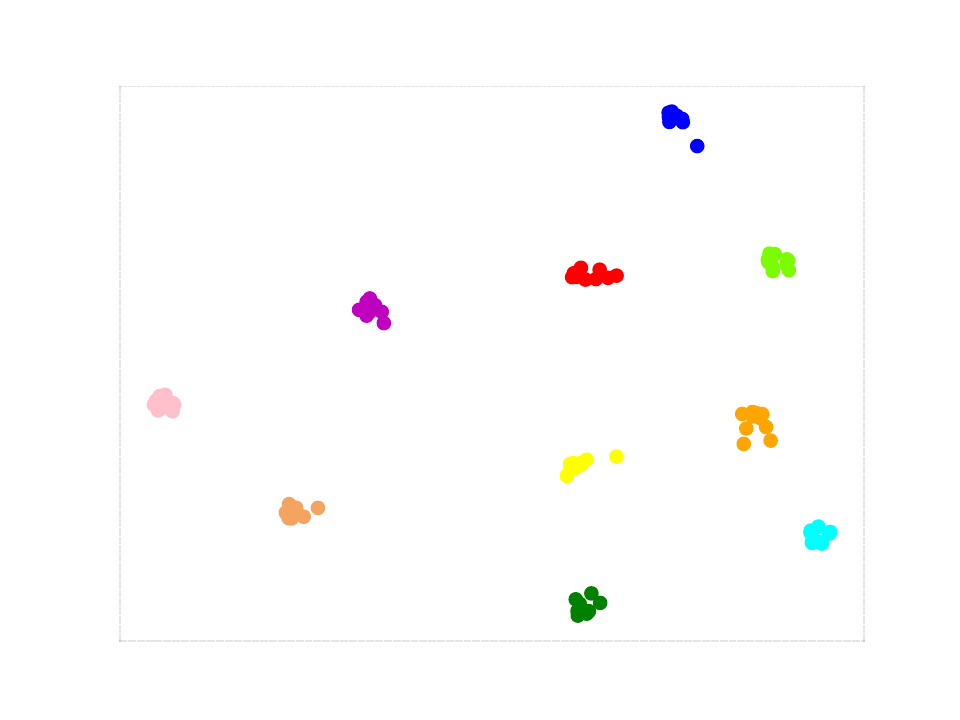}
            \caption{$\beta=1.2$}
        \end{subfigure}
        \hfill
        \begin{subfigure}{0.18\linewidth}
            \includegraphics[width=1\linewidth]{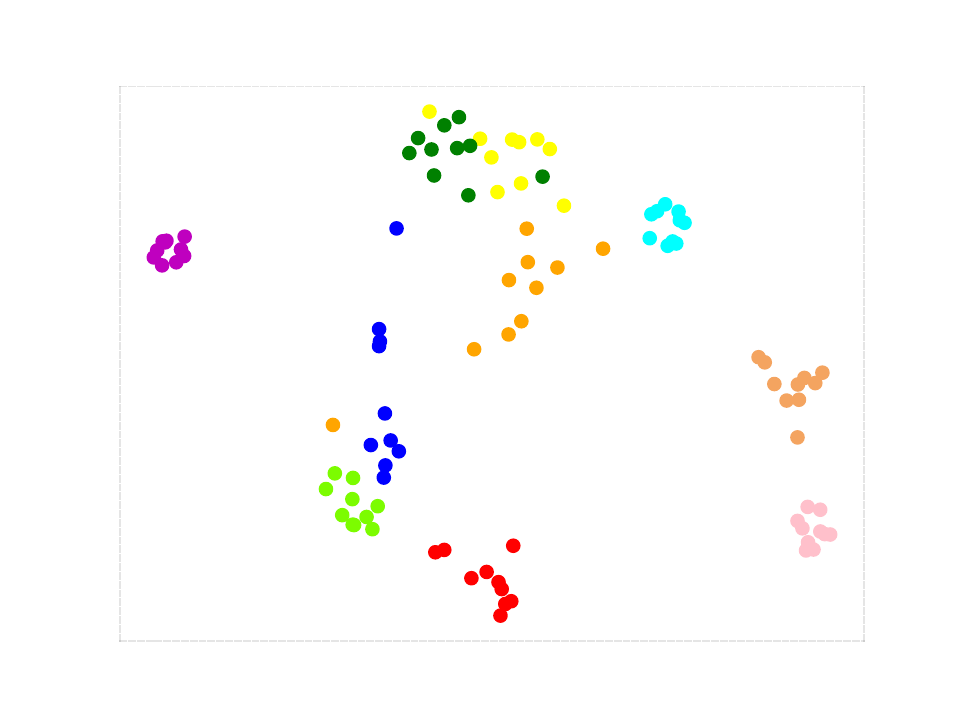}
            \caption{$\beta=1.3$}
        \end{subfigure}
        \hfill
        \begin{subfigure}{0.18\linewidth}
            \includegraphics[width=1\linewidth]{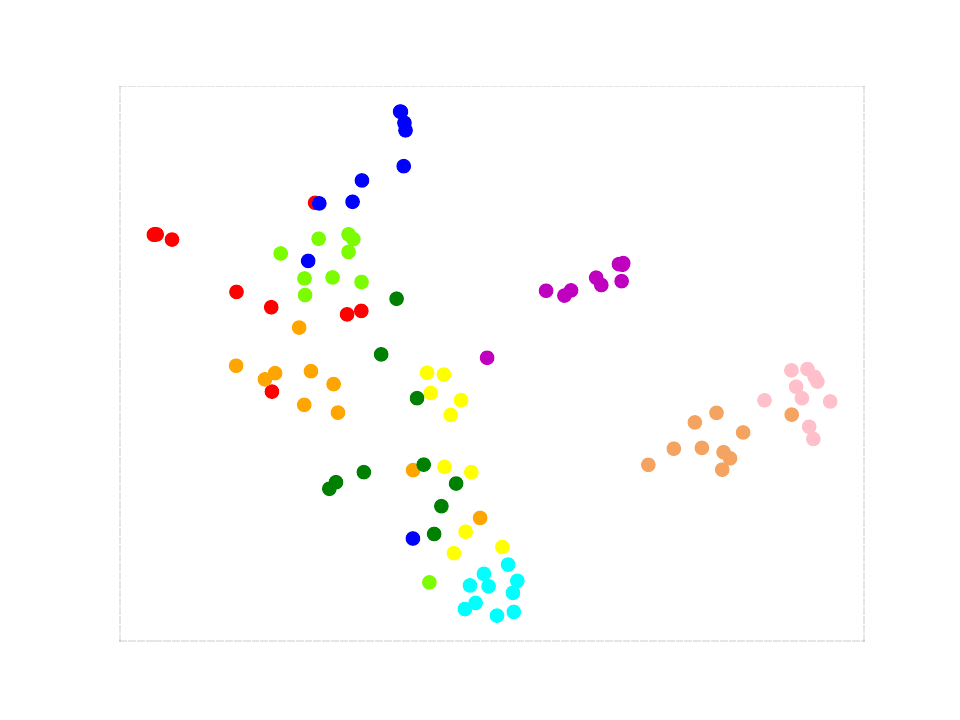}
            \caption{$\beta=1.5$}
        \end{subfigure} 
        \hfill
        \begin{subfigure}{0.18\linewidth}
            \includegraphics[width=1\linewidth]{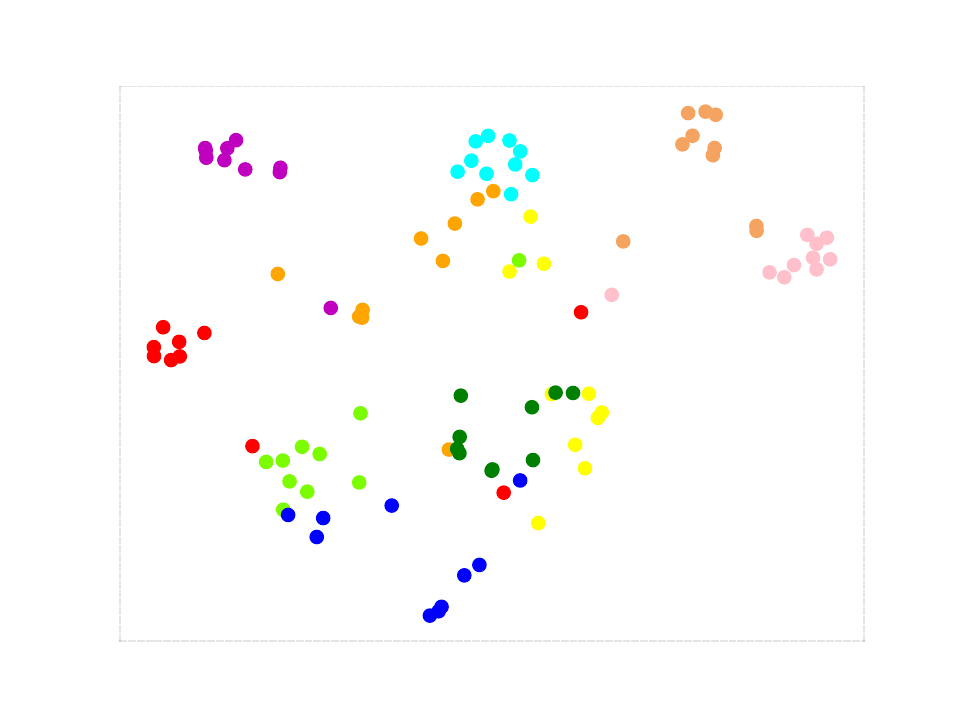}
            \caption{$\beta=2.0$}
        \end{subfigure}
        \hfill
      \caption{Visualization of different $\beta$ on CIFAR10 with IPC=10.}
      \label{fig:visbeta}
    \end{figure*}

\textbf{Analysis of cluster constraint threshold.}
The $\beta$ in \cref{eq:cluster} serves as the centralization threshold, where a larger $\beta$ indicates sample features are farther from the class feature center, while a smaller $\beta$  brings them closer. We conducted experiments with varying $\beta$ values, ranging from 0.0 to 2.0. To isolate the effects of $\beta$, we keep $\alpha$ in \cref{eq:cluster} constant and do not include our proposed covariance matching constraint.  The results  on CIFAR10 with IPC=10 are visualized in~\cref{fig:visbeta}, and the performance corresponding to each $\beta$ value is detailed in~\cref{tab:beta}.

\begin{table}[!tp]\small
    \centering
    \tabcolsep=3pt
    \caption{Cross-architecture testing on CIFAR10 with IPC=50.}
    \label{tab:cross50}
    \begin{tabular}{l*{5}{c}}
        \toprule
                 & ConvNet & AlexNet & VGG11 & ResNet18 \\
        \midrule
        DC~\cite{zhao2020dataset} & 53.9$\pm$0.5 & 28.8$\pm$0.7 & 38.8$\pm$1.1 & 20.9$\pm$1.0 \\
        CAFE~\cite{wang2022cafe} & 62.3$\pm$0.4 & 43.2$\pm$0.4 & 48.8$\pm$0.5 & 43.3$\pm$0.7 \\
        DSA~\cite{zhao2021dataset} & 60.6$\pm$0.5 & 53.7$\pm$0.6 & 51.4$\pm$1.0 & 47.8$\pm$0.9 \\
        DM~\cite{zhao2023dataset} & 63.0$\pm$0.4 & 60.1$\pm$0.5 & 57.4$\pm$0.8 & 52.9$\pm$0.4 \\
        KIP~\cite{nguyen2021dataset} & 56.9$\pm$0.4 & 53.2$\pm$1.6 & 53.2$\pm$0.5 & 47.6$\pm$0.8 \\
        DataDAM~\cite{sajedi2023datadam} & 67.0$\pm$0.4 & 63.9$\pm$0.9 & 64.8$\pm$0.5 & 60.2$\pm$0.7 \\
        \midrule
        \textbf{IDM+Ours} & \textbf{69.0$\pm$0.2} & \textbf{67.3$\pm$0.2} & \textbf{67.3$\pm$0.3} & \textbf{68.3$\pm$0.2} \\
        \bottomrule
    \end{tabular}
    \end{table}

     \begin{table*}[htp] \small
    \centering
    \tabcolsep=5pt
    \caption{Comparison of performance for different values of $\beta$ on CIFAR10 with IPC=10.}
    \label{tab:beta}
    \begin{tabular}{l *{10}{c}}
        \toprule
        $\beta$  & 0.0 & 0.1  & 0.7 & 0.9 & 1.0 & 1.1 & 1.2 & 1.3 & 1.5 & 2.0 \\
        \midrule
        Acc (\%)  & \textbf{52.2$\pm$0.2} & \textbf{52.2$\pm$0.1} & \textbf{52.2$\pm$0.1} & 51.5$\pm$0.3 & 51.1$\pm$0.2 & 49.9$\pm$0.2 & 49.2$\pm$0.2 & 49.1$\pm$0.1 & 48.9$\pm$0.3 & 48.9$\pm$0.1\\
        \bottomrule
    \end{tabular}
    \end{table*}
    
Our visualizations demonstrate that feature distribution per class becomes more dispersed with larger $\beta$ values and more concentrated with smaller $\beta$. Performance evaluation experiments reveal that optimal performance is achieved when $\beta$ is below 0.7. This suggests that smaller $\beta$ values, which lead to higher class discrimination, are advantageous for dataset distillation, enhancing performance without significantly compromising generalization.

\textbf{Effectiveness of each component.} To validate the effectiveness of the proposed two components, we perform ablation experiments. We test the performance impact of implementing the class centralization constraint and covariance matching constraint both separately and combined. \cref{tab:component} reveals that each constraint individually enhances DM’s performance by 3.7\% and 3.1\% on CIFAR10, and 1.4\% and 1.5\% on CIFAR100, respectively. Notably, the greatest performance improvement, 6.6\% on CIFAR10 and 2.5\% on CIFAR100, is achieved when both components are used concurrently.

    \begin{table}[htp]\small
    \centering
    \tabcolsep=4pt
    \caption{Ablation study on the proposed two components. Results are averaged over 5 runs.}
    \label{tab:component}
    \begin{tabular}{l c c c}
        \toprule
        Method & CIFAR10 & CIFAR100  \\
        \midrule
        Baseline & 48.5$\pm$0.2 & 29.7$\pm$0.1  \\
        + Class centralization    & 52.2$\pm$0.3 & 31.1$\pm$0.2  \\
        + Covariance matching   & 51.6$\pm$0.1 & 31.2$\pm$0.1  \\
        + Both   & \textbf{55.1$\pm$0.1}  & \textbf{32.2$\pm$0.3}  \\
        \bottomrule
    \end{tabular}
    \end{table}
    
\begin{figure}[htp]
      \centering
      \begin{subfigure}{0.48\linewidth}
        \includegraphics[width=1.0\linewidth]{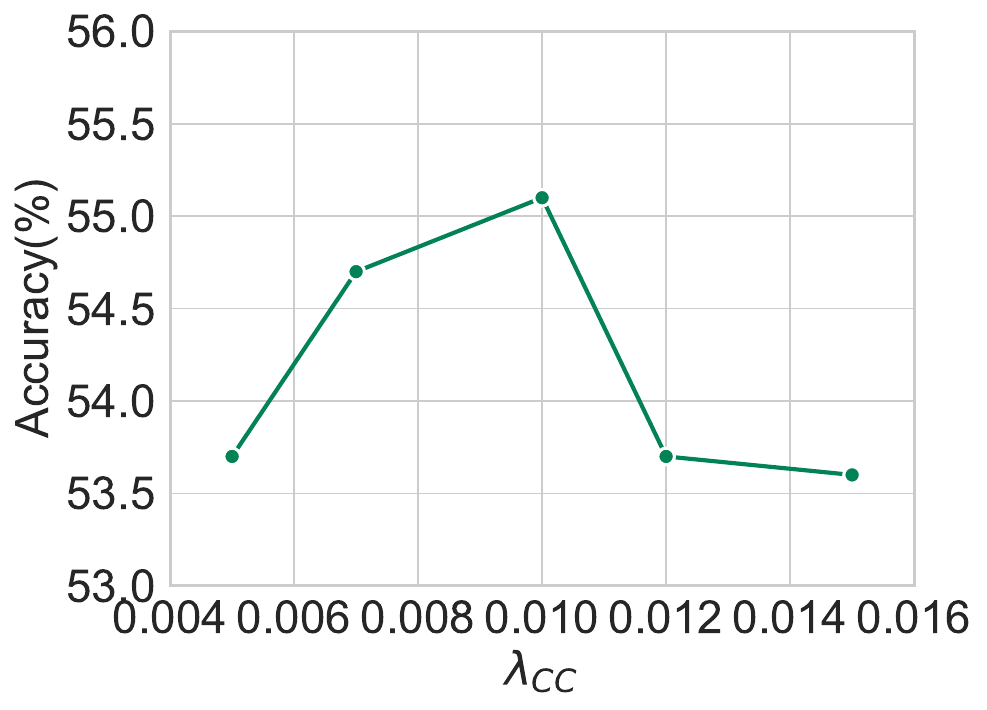}
        \caption{Ablation of $\lambda_{CC}$}
        \label{fig:lambda_clustering}
      \end{subfigure}
      \begin{subfigure}{0.48\linewidth}
        \includegraphics[width=1.0\linewidth]{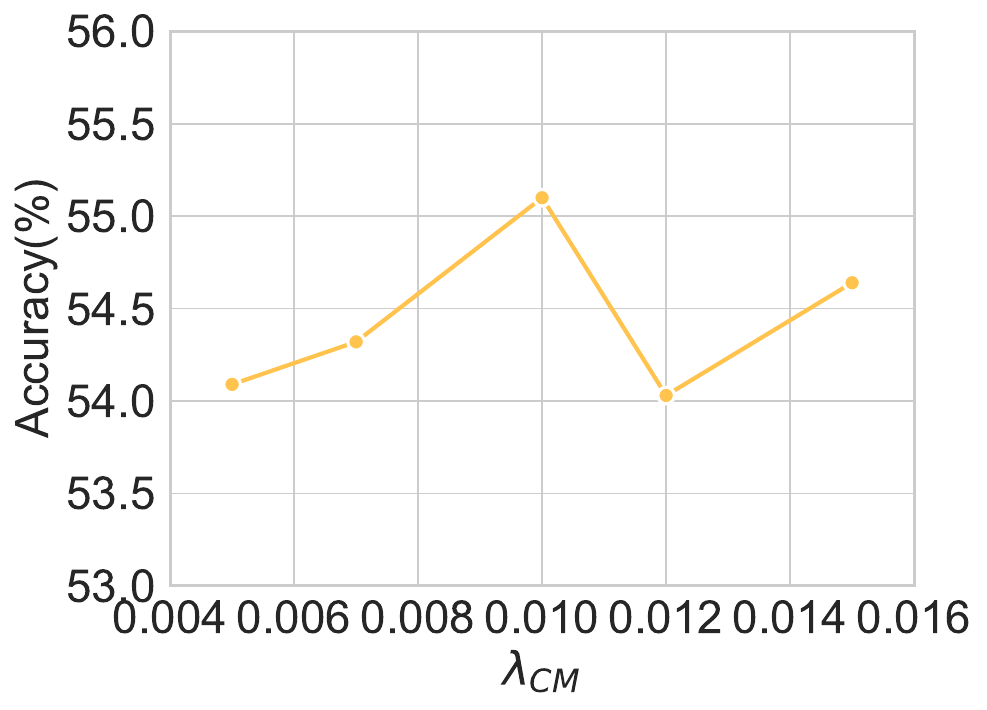}
        \caption{Ablation of $\lambda_{CM}$}
        \label{fig:lambda_local}
      \end{subfigure}
      \caption{Ablation study of the weighting parameter on CIFAR10.}
      \label{fig:lambda}
    \end{figure}

\textbf{Evaluation of weighting parameter $\lambda$.} We perform a sensitivity analysis focusing on the weighting parameters in~\cref{eq:overalldm}. The results, shown in~\cref{fig:lambda}, indicate that larger values of $\lambda_{CC}$ and $\lambda_{CM}$ correspond to a more pronounced impact on the optimization objective. Varying $\lambda_{CC}$ within a range of 0.01 to 0.09 resulted in a performance difference of 1.47\%, suggesting moderate sensitivity. Similarly, altering $\lambda_{CM}$ between 0.005 and 0.015 lead to a 1.1\%  performance variation, indicating comparable sensitivity levels. 

\begin{figure}[htp]
      \centering
      \begin{subfigure}{0.48\linewidth}
        \includegraphics[width=1.0\linewidth]{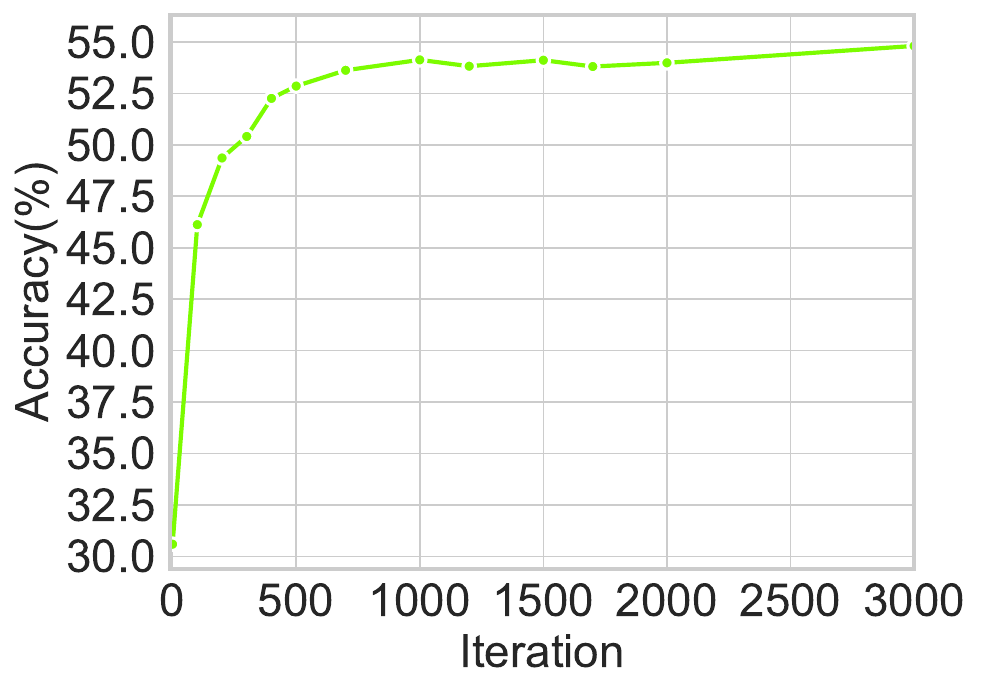}
        \caption{IPC=10}
        \label{fig:it10}
      \end{subfigure}
      \begin{subfigure}{0.48\linewidth}
        \includegraphics[width=1.0\linewidth]{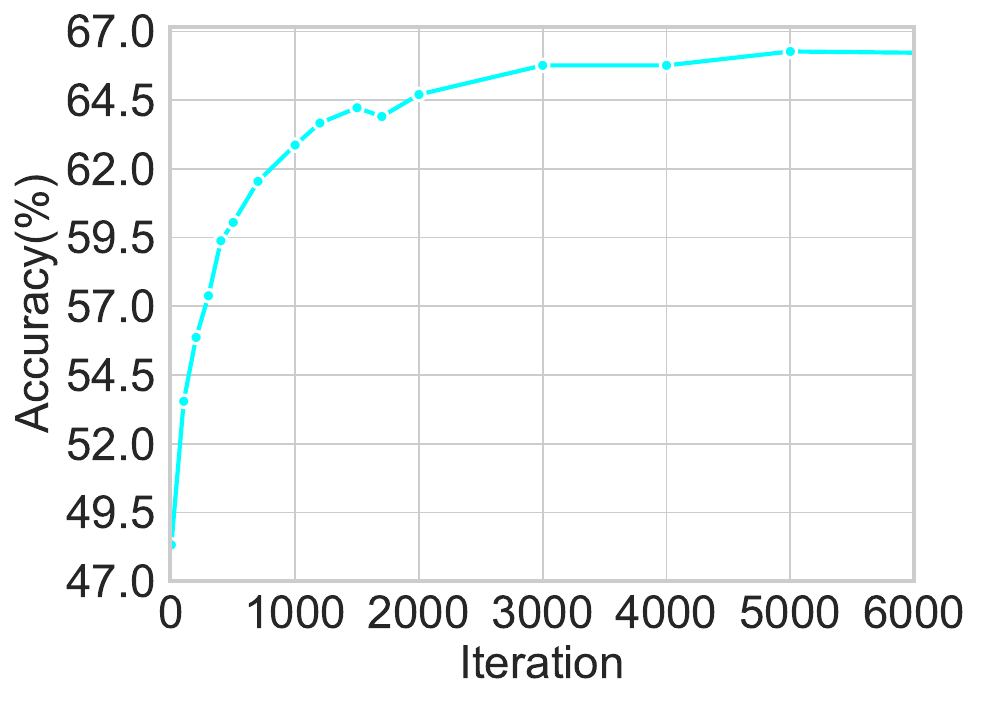}
        \caption{IPC=50}
        \label{fig:it50}
      \end{subfigure}
      \caption{Accuracy progression over iteration on CIFAR10.}
      \label{fig:it_ablation}
    \end{figure}

\textbf{Number of iterations required for convergence.} Contrary to previous methods which typically require around 20,000 training iterations, our method achieves convergence with significantly fewer iterations. As demonstrated in~\cref{fig:it_ablation}, less than 2,000 iterations are needed for IPC=10 to attain peak performance. For IPC=50, optimal performance is essentially reached at about 3,000 iterations. In addition to this, we find that accuracy grows quickly in the early stages of training, and if the training time requirement is high while the performance requirement is relatively low, we can consider using early stopping training methods.

\begin{figure}[htp]
       \centering
        \includegraphics[width=0.65\linewidth]{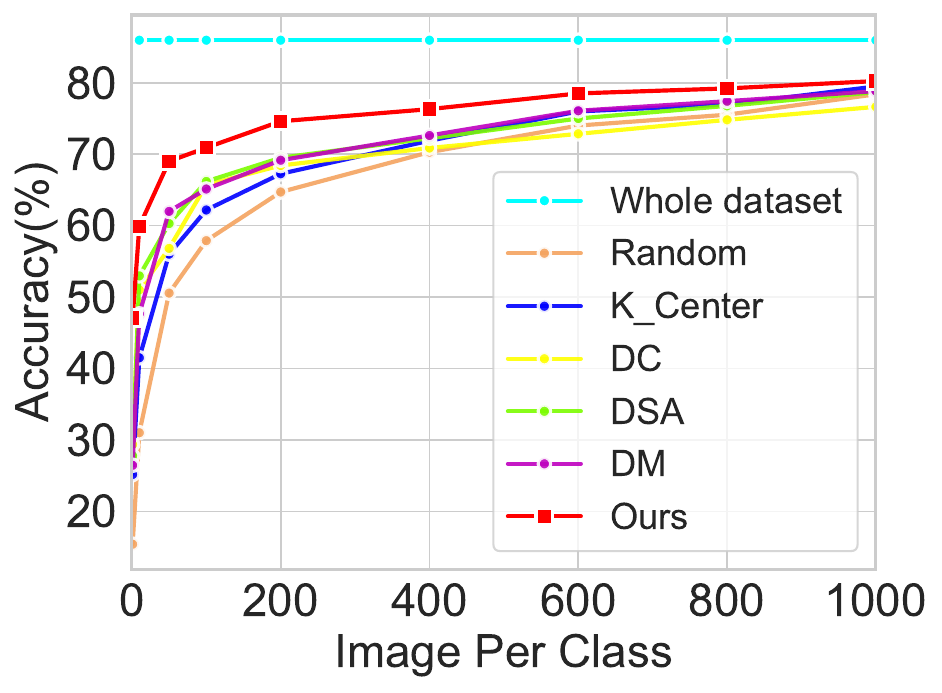}
        \caption{Performance comparison at different compression ratios on CIFAR10.}
        \label{fig:largeipc}
        \end{figure}

\textbf{Different compression ratios.}
Previous dataset distillation studies have primarily focused on improving performance at very small compression ratios, such as 0.2\% (IPC=10 on CIFAR10) and 1\% (IPC=50 on CIFAR10). However, an equally important consideration is determining the necessary compression ratio to retain performance comparable to the full dataset. To explore this, we conduct experiments at higher IPC values, including IPC=100, 200, 400, 600, 800, and 1000, and compared our method's performance with previous methods, as shown in~\cref{fig:largeipc,tab:largeipc}, with most results sourced from DC-BENCH~\cite{cui2022dc}.

Our method significantly outperforms others like DC~\cite{zhao2020dataset}, DSA~\cite{zhao2021dataset}, and  DM~\cite{zhao2023dataset} at these varied IPC levels. Notably, at IPC=1000 (a 20\% compression ratio), our method achieves 93.3\% of the full dataset's performance. However, it's observed that as the compression ratio increases, the performance gap between all methods, including classical and even random methods, narrows. This suggests that current dataset distillation approaches are more effective at smaller compression ratios. In future research, it would be valuable to explore dataset distillation methods suitable for larger compression ratios, aiming for an optimal balance between performance and data reduction.

        \begin{table}[!t]\small
        \centering
        \tabcolsep=5pt
        \caption{Accuracy (\%) with different compression ratios on CIFAR10.}
        \label{tab:largeipc}
        \begin{tabular}{c *{6}{c}}
            \toprule
            IPC & Random & K-Center & DC & DSA & DM & \textbf{Ours} \\
            \midrule
            1     & 15.40 & 25.16 & 29.34 & 27.76 & 26.45 & \textbf{47.11}\\
            10    & 31.00 & 41.49 & 50.99 & 52.96 & 47.64 & \textbf{59.92}\\
            50    & 50.55 & 56.00 & 56.81 & 60.28 & 61.99 & \textbf{69.01}\\
            100   & 57.89 & 62.18 & 65.70 & 66.18 & 65.12 & \textbf{70.46}\\
            200   & 64.70 & 67.25 & 68.41 & 69.49 & 69.15 & \textbf{74.15}\\
            400   & 70.28 & 71.88 & 70.86 & 72.22 & 72.61 & \textbf{76.35}\\
            600   & 74.00 & 75.98 & 72.84 & 74.99 & 76.07 & \textbf{78.38}\\
            800   & 75.52 & 76.94 & 74.80 & 76.76 & 77.41 & \textbf{79.21}\\
            1000  & 78.38 & 79.47 & 76.62 & 78.68 & 78.83 & \textbf{80.31}\\
            \bottomrule
        \end{tabular}
        \end{table}
        
\subsection{Applications}

\begin{figure}[!tp]
      \centering
      \begin{subfigure}{0.48\linewidth}
        \includegraphics[width=1.0\linewidth]{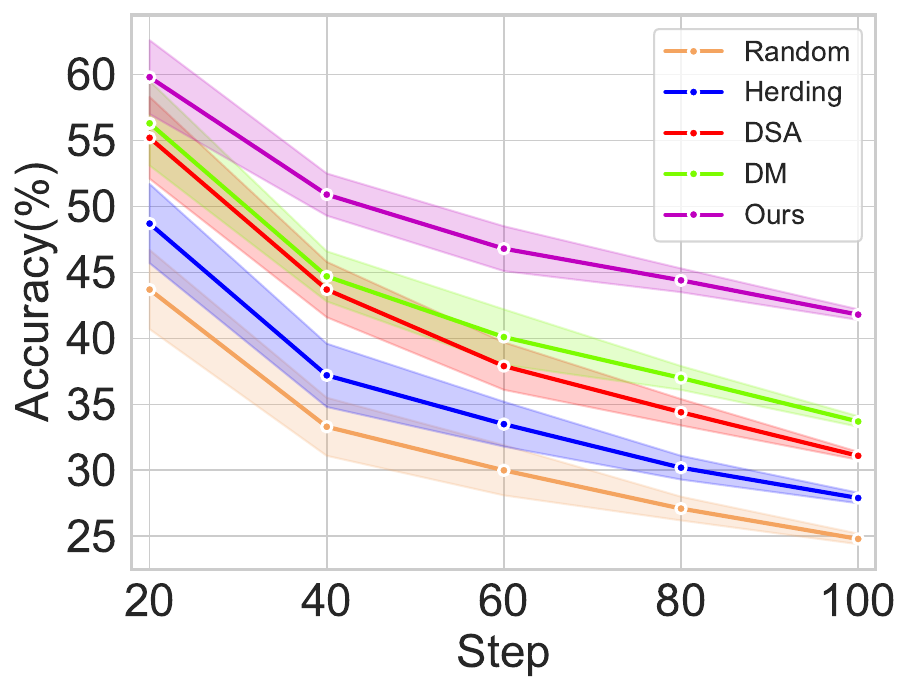}
        \caption{Step=5}
        \label{fig:cl5}
      \end{subfigure}
      \begin{subfigure}{0.48\linewidth} 
        \includegraphics[width=1.0\linewidth]{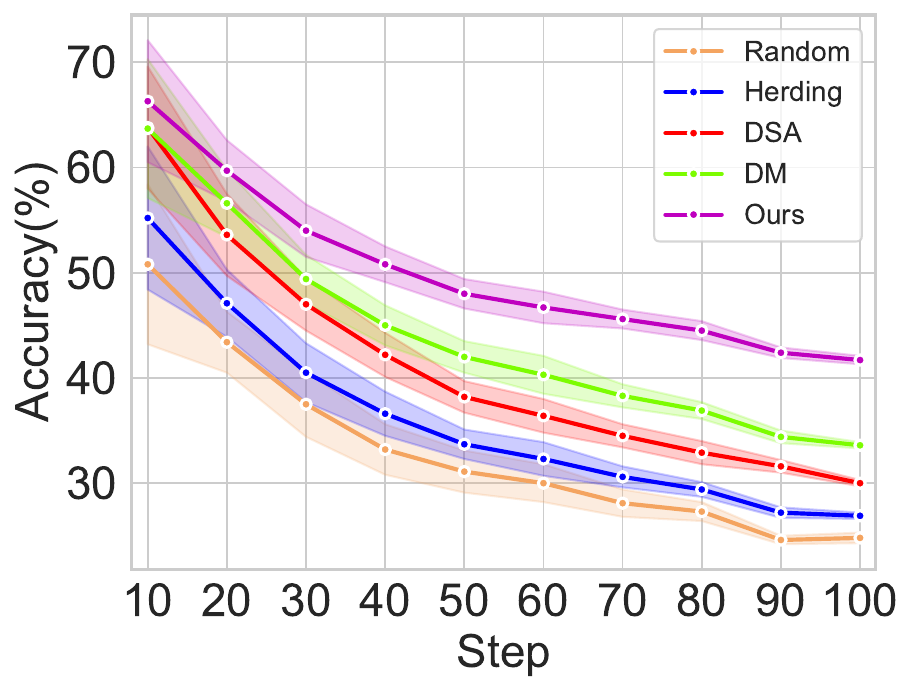}
        \caption{Step=10}
        \label{fig:cl10}
      \end{subfigure}
      \caption{Continual learning on CIFAR100.}
      \label{fig:cl}
    \end{figure}

\textbf{Continual learning.} The goal of continual learning is to develop models capable of adapting to new tasks while minimizing the forgetting of previously learned tasks. 
Dataset distillation, with its ability to maintain performance with a minimal number of samples, has potential applications in continual learning. Following the setup from the prior work~\cite{zhao2023dataset}, we limit the buffer size to 20 images per category. Experiments are conducted in both 5-step (20 classes per task) and 10-step (10 classes per task) scenarios. For each step, three networks are randomly initialized, trained on the synthetic dataset from the buffer, and evaluated on a test set of seen classes to determine the mean performance. Five class orders are randomly generated for these experiments, and the mean and variance are calculated. The results, depicted in~\cref{fig:cl}, show that our method notably outperforms DM~\cite{zhao2023dataset}, DSA~\cite{zhao2021dataset}, and Herding~\cite{welling2009herding}, with the performance gap widening as the number of steps increases. 

\subsection{Visualization}
\begin{figure}[t]
      \centering
        \begin{subfigure}{0.46\linewidth}
        \includegraphics[width=1\linewidth]{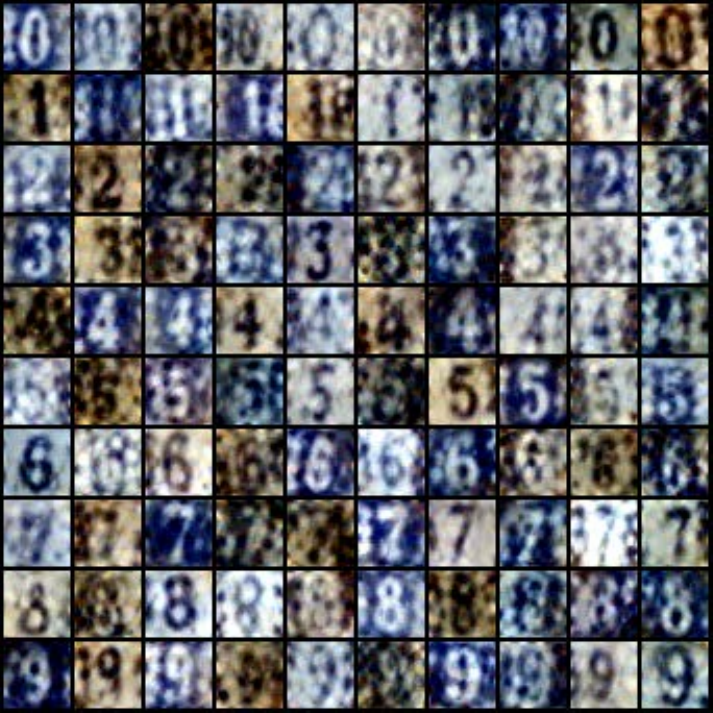}
        \caption{SVHN}
        \label{fig:svhn_vis}
      \end{subfigure}
      \begin{subfigure}{0.46\linewidth}
        \includegraphics[width=1\linewidth]{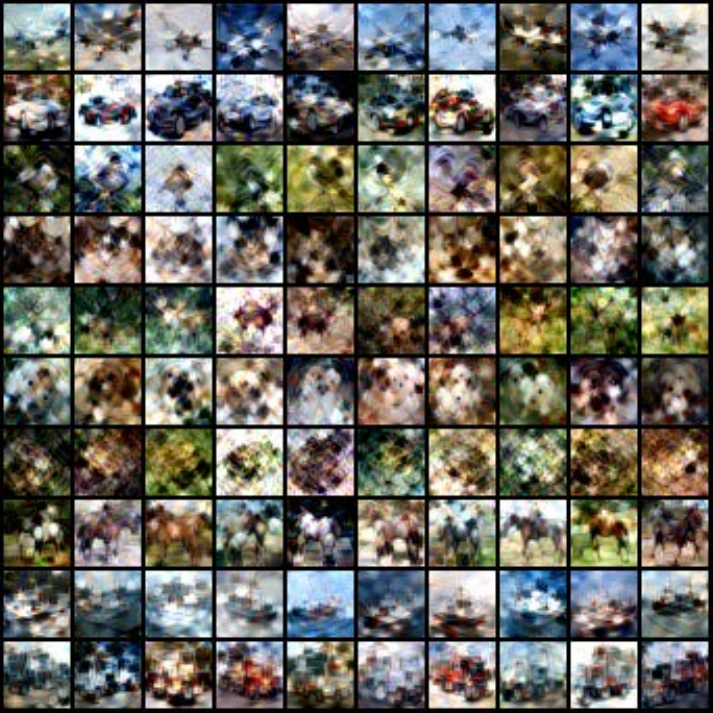}
        \caption{CIFAR10}
        \label{fig:cifar10_vis}
      \end{subfigure}
      \begin{subfigure}{0.46\linewidth}
        \includegraphics[width=1\linewidth]{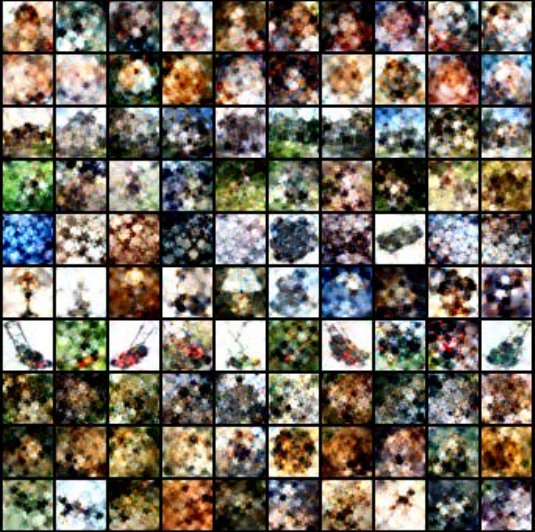}
        \caption{CIFAR100 (Partial)}
        \label{fig:cifar100_vis}
      \end{subfigure}
      \begin{subfigure}{0.46\linewidth}
        \includegraphics[width=1\linewidth]{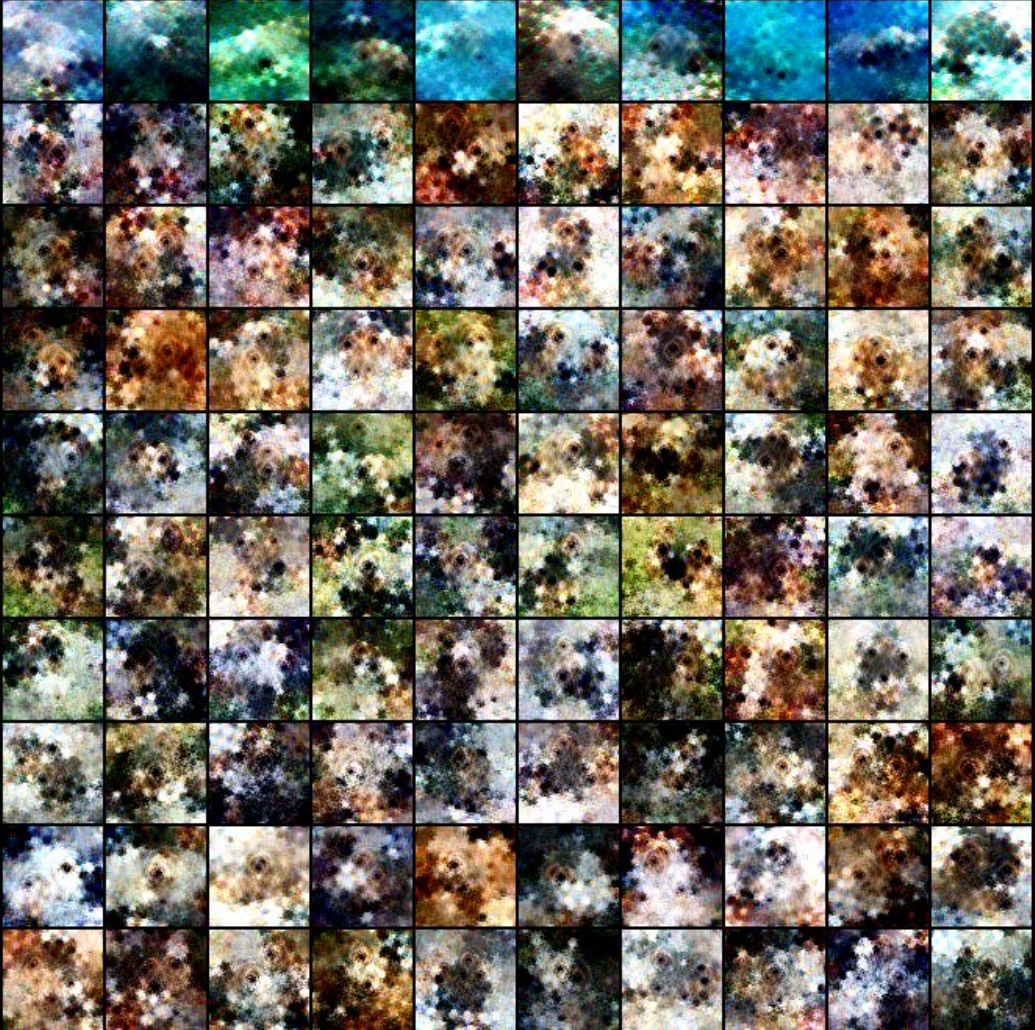}
        \caption{TinyImageNet (Partial)}
        \label{fig:tiny_vis}
      \end{subfigure}
      \caption{Visualization of synthetic images.}
      \label{fig:vis}
    \end{figure}

The synthetic datasets, initialized randomly from real datasets, are visualized in~\cref{fig:vis}. These include synthesized datasets from SVHN, CIFAR10/100, and Tiny-ImageNet at IPC=10. For datasets with smaller resolutions like SVHN and CIFAR10/100, the majority of high-frequency information is preserved, making them more recognizable to the human eye. In contrast, for higher resolution datasets such as Tiny-ImageNet, the synthetic dataset's visualization appears much different from the real dataset, becoming less distinguishable visually. Differential visualization results appeared between low-resolution data sets and high-resolution data sets, suggesting that in the future we may need to design different dataset distillation methods for datasets of different scales.

\section{Conclusion}
Previous distribution matching-based methods for dataset distillation face two primary challenges: insufficient class discrimination and incomplete distribution matching. To address these, we introduced the class centralization constraint and the covariance matching constraint, focusing on improving both inter-sample and inter-feature relations. The class centralization constraint improves class discrimination by clustering samples closer to their class centers, while the covariance matching constraint aligns inter-feature relationships between real and synthetic datasets. Our method, tested across various resolutions, has demonstrated significant superiority over previous methods and excelled in cross-architecture scenarios. Additionally, we explored larger compression ratios to determine the necessary compression ratio for maintaining performance comparable to the full dataset. 

\noindent \textbf{Acknowledgements.} This work is supported in part by the National Natural Science Foundation of China (62106100, 62192783, 62276128, 62106282), Science and Technology Innovation 2030 New Generation Artificial Intelligence Major Project (2021ZD0113303), Jiangsu Natural Science Foundation (BK20221441), Young Elite Scientists Sponsorship Program by CAST (2023QNRC001), the Collaborative Innovation Center of Novel Software Technology and Industrialization, Beijing Nova Program (20220484139).

{
    \small
    \bibliographystyle{ieeenat_fullname}
    \bibliography{main}
}

\end{document}